\documentclass[letterpaper, 10 pt, conference]{ieeeconf}  % Comment this line out if you need a4paper

\IEEEoverridecommandlockouts                              % This command is only needed if
\usepackage[maxbibnames=6]{biblatex}
\usepackage{url, amsfonts, amsmath, hyperref, subcaption}
\usepackage{dblfloatfix}    % To enable figures at the bottom of page
\usepackage{graphicx}

\usepackage{gensymb}
\usepackage{multirow}
\usepackage{MnSymbol}
\usepackage{wasysym}
\usepackage{float}
\usepackage[x11names]{xcolor}
\usepackage{soul}
\usepackage[normalem]{ulem}

\usepackage[labelsep=period,skip=1pt]{caption}
\usepackage{dblfloatfix} 
\usepackage{etoolbox}
\newtoggle{vtwo}
\newtoggle{vtwo_hl}
\newtoggle{vfinal}
\newtoggle{vfinal_hl}

\togglefalse{vtwo}
\togglefalse{vtwo_hl}
\togglefalse{vfinal}
\togglefalse{vfinal_hl}

\newcommand{\secondversion}[1]{%
\iftoggle{vtwo}{ \textcolor{red}{\sout{#1}}}
{}
}

\newcommand{\hlsecondversion}[1]{%
\iftoggle{vtwo_hl}{\hl{#1}}
{#1}
}

\newcommand{\finalversion}[1]{%
\iftoggle{vfinal}{\textcolor{red}{\sout{#1}}}
{}
}

\newcommand{\hlfinalversion}[1]{%
\iftoggle{vfinal_hl}{\hl{#1}}
{#1}
}

\usepackage{array, booktabs, makecell} 
\usepackage{siunitx, mhchem}

\usepackage[letterpaper, left=0.667in, right=0.667in, bottom=0.597in, top=0.833in]{geometry}

\title{
Soft Lattice Modules that Behave Independently and Collectively
}
\author{
\authorblockN{Luyang Zhao\authorrefmark{1}, Yijia Wu\authorrefmark{1}, Julien Blanchet\authorrefmark{1}, Maxine Perroni-Scharf\authorrefmark{3}, \\ Xiaonan Huang\authorrefmark{2}, Joran Booth\authorrefmark{2}, Rebecca Kramer-Bottiglio\authorrefmark{2}, Devin Balkcom\authorrefmark{1}}
\authorblockA{Dartmouth College\authorrefmark{1}, Yale University\authorrefmark{2}, Princeton University\authorrefmark{3}}
}

\begin{document}
\maketitle
%\institute{Dartmouth College}

\begin{abstract}
    
    % This paper explores the first step using a symmetric designed modular tensegrity robot to 
    % achieve single module locomotion, attachments between single modules, multiple modules locomotion,
    % and bi-directional non-prehensile manipulations. 
    % Inside, we explored the structure typologies emerging from tensegrity assemblies and tetrahedral molecular that can support both manipulation and locomotion, where the structure of tetrahedral molecular is referred and a symmetric modification is made to make sure each robot is gravitational stable. 
    
    % djb polishing pass on abstract. 
    Natural systems integrate the work of many sub-units (cells) toward a large-scale unified goal (morphological and behavioral), which can counteract the effects of unexpected experiences, damage, or simply changes in tasks demands. In this paper, we exploit the opportunities presented by soft, modular, and tensegrity robots to introduce soft lattice modules that parallel the sub-units seen in biological systems. The soft lattice modules are comprised of 3D printed plastic ``skeletons", linear contracting shape memory alloy spring actuators, and permanent magnets that enable adhesion between modules. The soft lattice modules are capable of independent locomotion, and can also join with other modules to achieve collective, self-assembled, larger scale tasks such as collective locomotion and moving an object across the surface of the lattice assembly. This work represents a preliminary step toward soft modular systems capable of independent and collective behaviors, and provide a platform for future studies on distributed control. 
    
    %This paper presents a design for a self-assembling soft modular lattice. Using tethered control, the modules achieve an end-to-end scenario in which several modules assemble into a connected mesh, locomote collectively, and perform peristaltic manipulation of a ball atop the mesh surface. The robot skeletons are 3D printed using a flexible material, actuated using shape-memory-alloy springs, and connected together with permanent magnets. 
\end{abstract}

\section{Introduction}
\label{sec:introduction}

% What is the paper about?
% % Note: the abstract gives a overview of what this paper is about, so it's ok if we start the introduction by providing some context.
% \hl{todo: what this paper is about, in brief.}
% In this paper, we explored how to use the property of tensegrity to explore modular tensegrity robots with simple and efficient structure to achieve locomotion, attachments, and all-direction non-prehensile manipulation. See Fig. \ref{fig:system_overview} for an overview.

% \begin{figure*}[htb]
%     \centering
    
%     \begin{subfigure}{.23\linewidth}
%       \includegraphics[width=\linewidth]{images/single_module.png}
%       \caption{Single Module}
%     \end{subfigure}
%     \begin{subfigure}{.23\linewidth}
%       \includegraphics[width=\linewidth]{images/test.png}
%       \caption{3-Module Assembly}
%     \end{subfigure}
%     \begin{subfigure}{.23\linewidth}
%       \includegraphics[width=\linewidth]{images/Screen Shot 2021-08-21 at 11.49.15 AM.png}
%       \caption{Lattice Self-Assembly}
%     \end{subfigure}
%     \begin{subfigure}{.23\linewidth}
%       \includegraphics[width=\linewidth]{images/manipulation_3d.pdf}
%       \caption{Peristaltic Manipulation}
%     \end{subfigure}
%     \caption{System Overview}
%     \label{fig:system_overview}
% \end{figure*}

% djb working on polishing pass
% Why is it important?

This paper presents soft\secondversion{lattice}modules that can attach to one another to form larger, more capable lattices. \hlsecondversion{Different shapes of lattice assemblies support different tasks, and we demonstrate configurations performing locomotion, peristaltic surface manipulation and prehensile gripping.}

We are particularly interested in exploring the intersection of work on soft robots and modular robots. While modular robot systems collectively reconfigure themselves to allow manipulation or locomotion, flexible lattices allow deformation as well as reconfiguration, motivating the exploration of new behaviors; for example, in this paper, we explore whole-body manipulation of a ball along the surface of the lattice.
%\textcolor{red}{\sout{-- for example}}, In this paper, we explore whole-body manipulation of a ball along the surface of the lattice.
%that builds off prior work in soft robotics, self-reconfiguring robots, and peristaltic manipulation. 
%Our goal is to leverage the advantages of a flexible robot body in a multi-robot system that can locomote, self-assemble in to a lattice structure, and manipulate objects atop the lattice surface. 
Eventually, we imagine that such a system could be used as a rapidly deployable, flexible structure with applications in disaster relief and construction, capable of serving as a 2D conveyance for materials or as active scaffolding for construction in hostile environments.

% Can serve to transport materials and partially assembled components along the surface while also serving as a foundation. Could potentially work underwater and help facilitate automated construction in that dynamic environment.

Soft robots constructed of flexible materials offer the possibility of simple mechanical designs, safe operation in the presence of humans and delicate objects, excellent resistance to mechanical impact, and compliance to adapt to uneven terrain~\cite{lee2017soft}. Owing to high force-to-mass ratio, high work density,\secondversion{fast actuation response}and the capability of being actuated by light on-board electronics \cite{huang2020shape}, many soft robots are actuated using spring-memory alloy (SMA) variable length tendons \cite{lee2017soft}, which can be combined with a flexible body to enable limb-like bending. 

Collective behavior to perform complex tasks\secondversion{that would be impossible for a single module to perform alone}has garnered interest in recent years \cite{Yim2002ModularRobots}. One thrust of prior research has been on small rigid-bodied units capable of rearranging into various chains or circular configurations in order to achieve novel locomotive capabilities~\cite{mondada2004swarm, Murata2002M_TRAN, Parker2016}. Another thrust has been self-assembly and reconfiguration of lattice-based systems, in which robot modules are connected in a grid-like pattern \cite{Parker2016}.

\begin{figure}[bth!]
    \centering
    \begin{subfigure}{.39\linewidth}
     \centering
      \includegraphics[height=3cm]{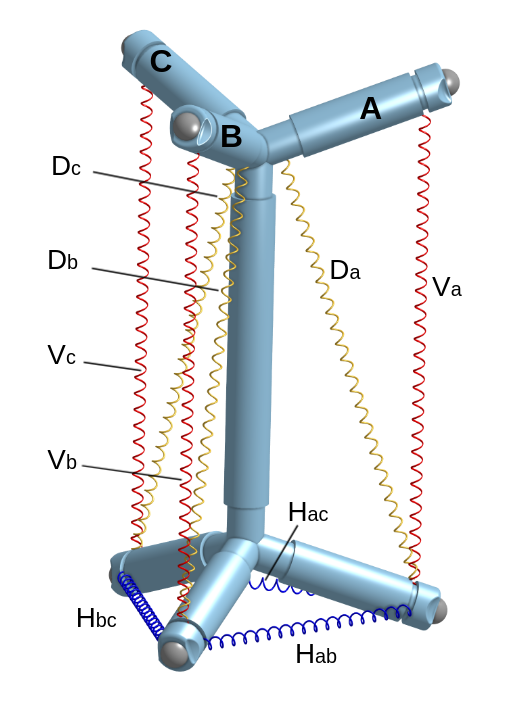} 
      \caption{Single module}
      \label{fig:system_overview_singlemodule}
    \end{subfigure}
    \begin{subfigure}{.58\linewidth}
    \centering
      \includegraphics[height=3cm]{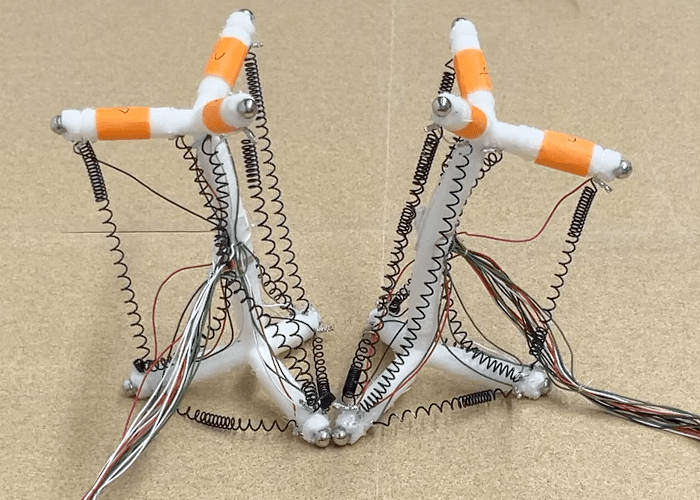}
      \caption{Module attachment}
      \label{fig:system_overview_attachment}
    \end{subfigure}
    \begin{subfigure}{\linewidth}
    \centering
\begin{subfigure}{.24\linewidth}
 \renewcommand\thesubfigure{\alph{subfigure}1}
  \centering
  \includegraphics[width=\linewidth]{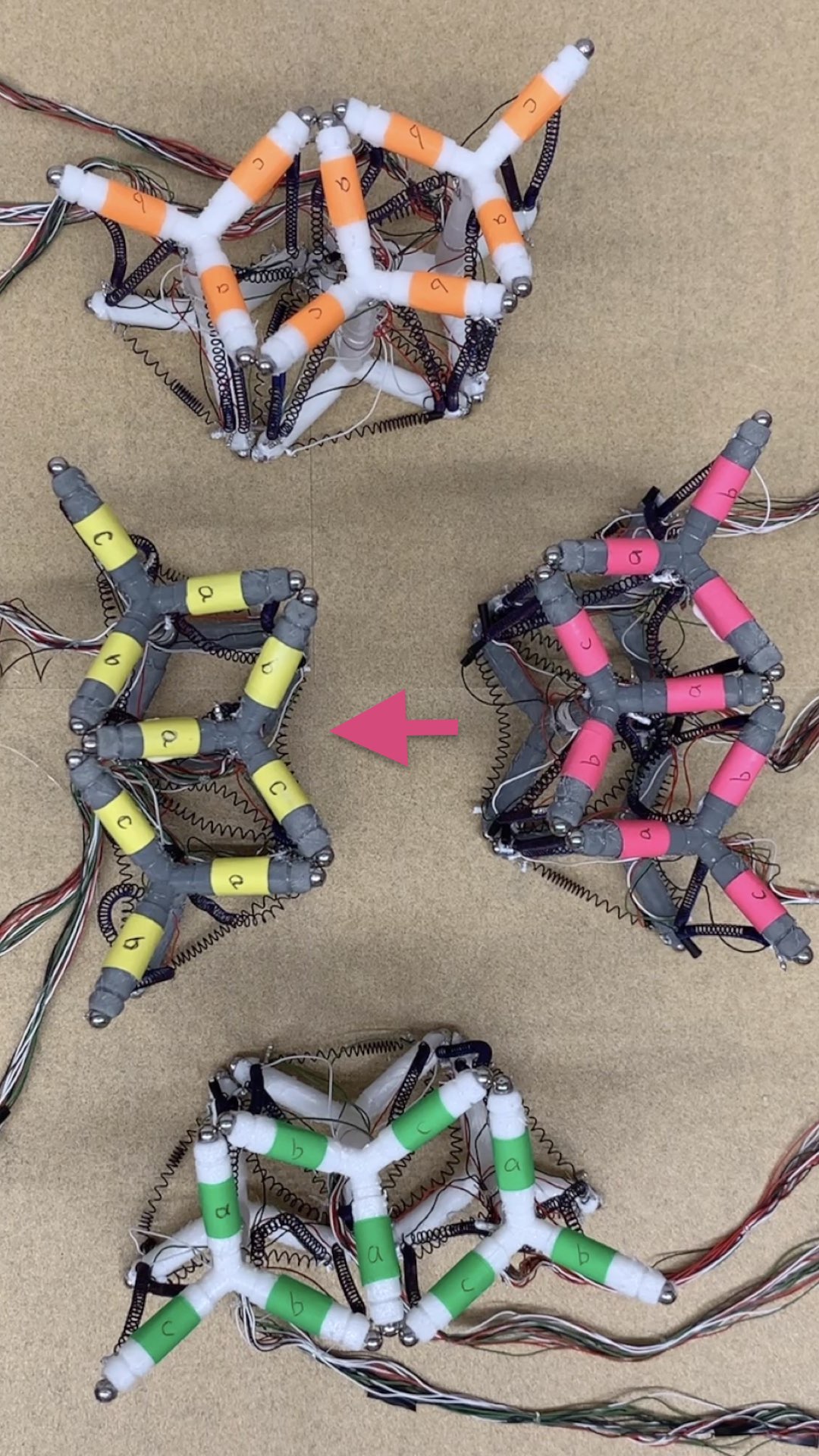}
  \caption{} 
  \label{fig:c1} 
\end{subfigure}
\begin{subfigure}{.24\linewidth}
\addtocounter{subfigure}{-1}
 \renewcommand\thesubfigure{\alph{subfigure}2}
  \centering
  \includegraphics[width=\linewidth]{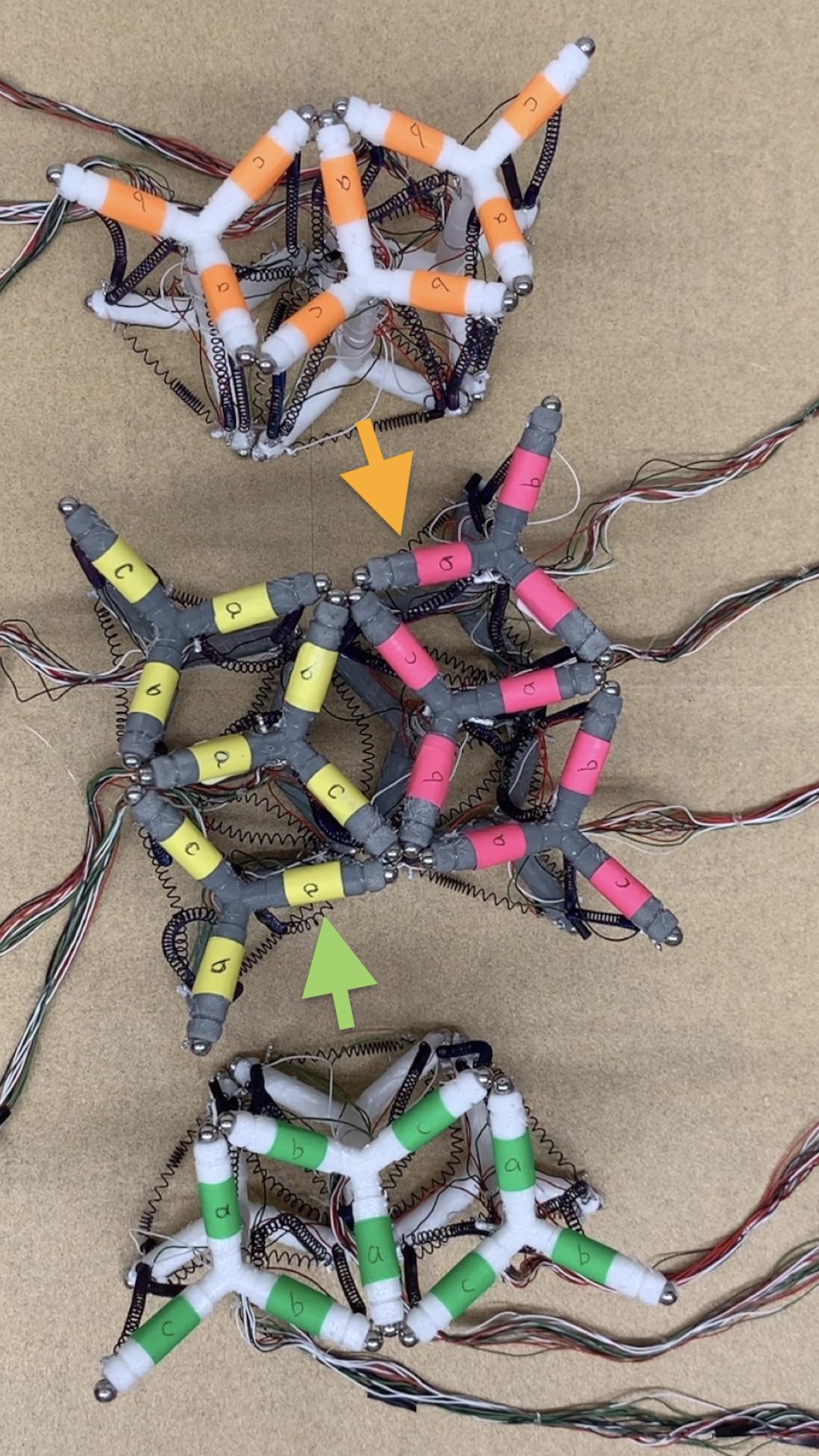}
   \caption{} 
   \label{fig:c2} 
    \end{subfigure}
    \begin{subfigure}{.24\linewidth}
    \addtocounter{subfigure}{-1}
     \renewcommand\thesubfigure{\alph{subfigure}3}
      \centering
      \includegraphics[width=\linewidth]{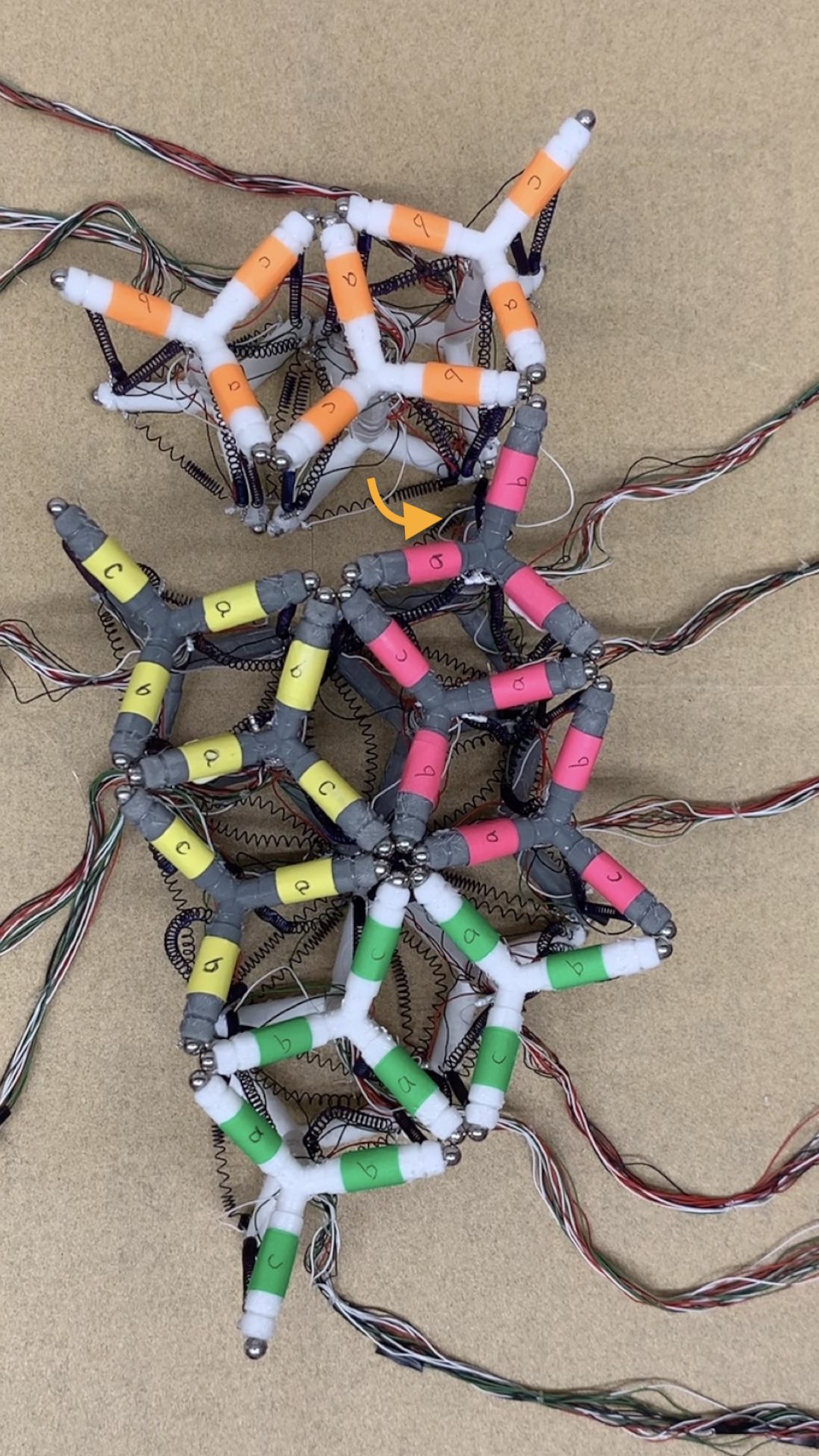}
      \caption{} 
      \label{fig:c3} 
    \end{subfigure}
    \begin{subfigure}{.24\linewidth}
    \addtocounter{subfigure}{-1}
     \renewcommand\thesubfigure{\alph{subfigure}4}
      \centering
      \includegraphics[width=\linewidth]{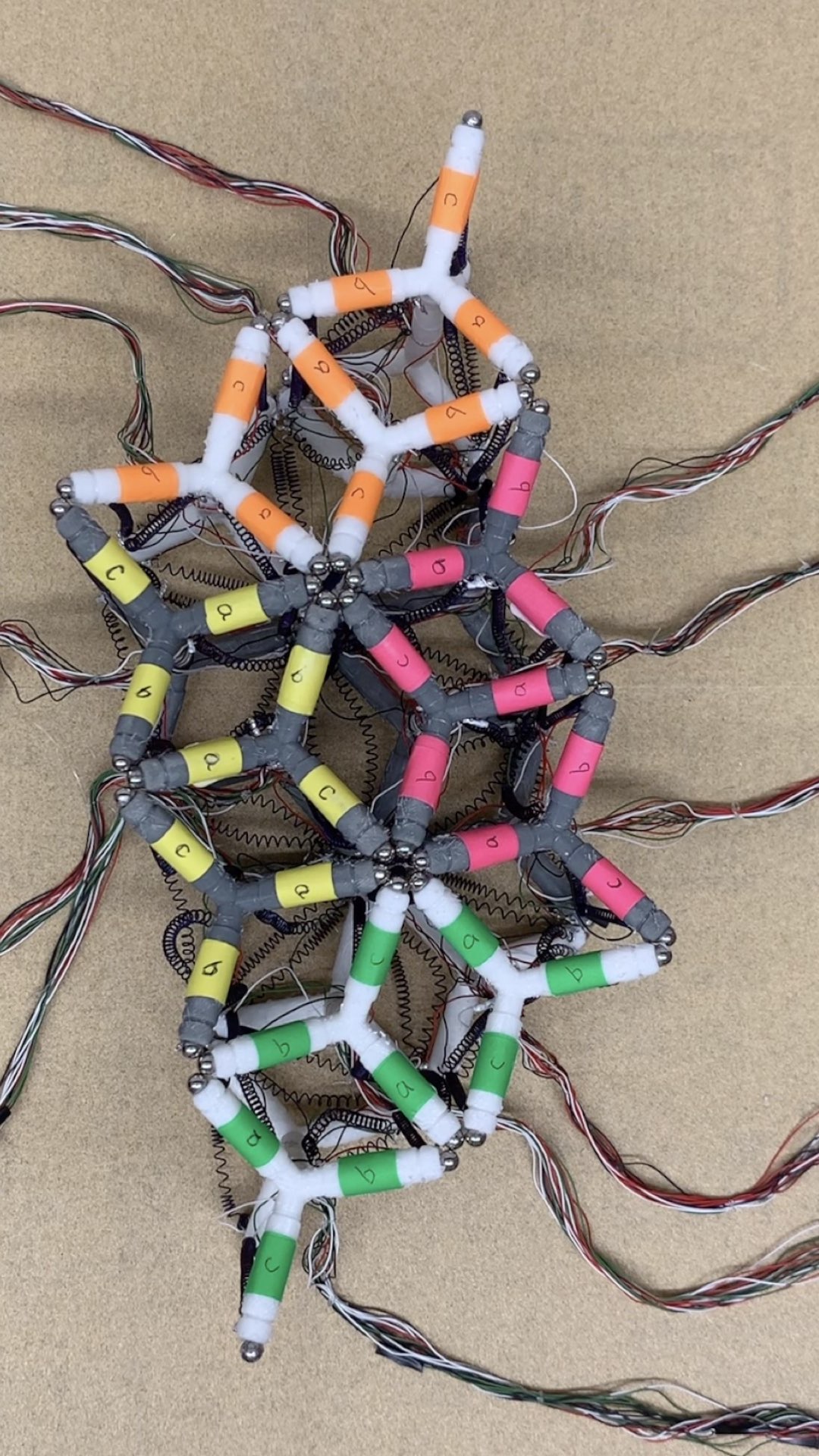}
      \caption{} 
      \label{fig:c4} 
    \end{subfigure}
    \addtocounter{subfigure}{-1}
      \caption{Lattice self-Assembly}
      \label{fig:system_overview_lattice_assembly}
    \end{subfigure}
    \begin{subfigure}{\linewidth}
      \includegraphics[width=\linewidth]{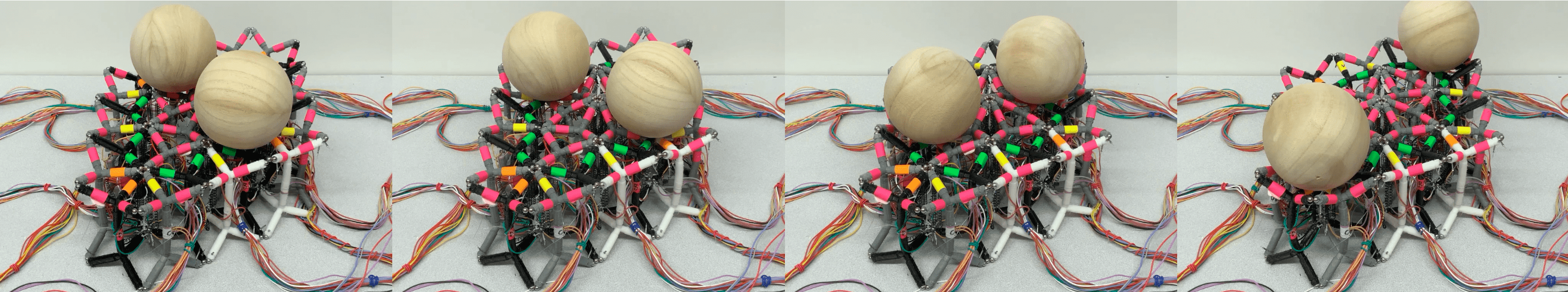}
      \caption{Manipulating two balls}
      \label{fig:system_overview_manipulation}
    \end{subfigure}
    % \begin{subfigure}{\linewidth}
    %   \includegraphics[width=\linewidth]{images/gripper2.png}
    %   \caption{\hl{Module-base gripper}}
    %   \label{fig:system_overview_gripper}
    % \end{subfigure}
        \begin{subfigure}{\linewidth}
    \centering
\begin{subfigure}{.16\linewidth}
 \renewcommand\thesubfigure{\alph{subfigure}1}
  \centering
  \includegraphics[width=\linewidth]{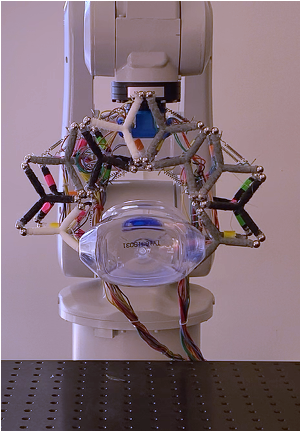}
  \caption{} 
  \label{fig:g1} 
\end{subfigure}%
\begin{subfigure}{.16\linewidth}
\addtocounter{subfigure}{-1}
 \renewcommand\thesubfigure{\alph{subfigure}2}
  \centering
  \includegraphics[width=\linewidth]{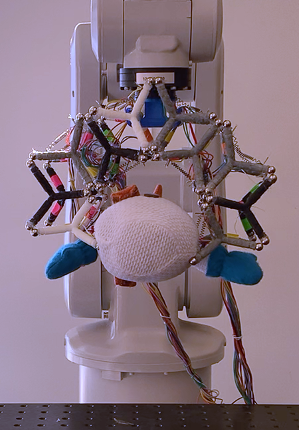}
   \caption{} 
   \label{fig:g2} 
    \end{subfigure}%
    \begin{subfigure}{.16\linewidth}
    \addtocounter{subfigure}{-1}
     \renewcommand\thesubfigure{\alph{subfigure}3}
      \centering
      \includegraphics[width=\linewidth]{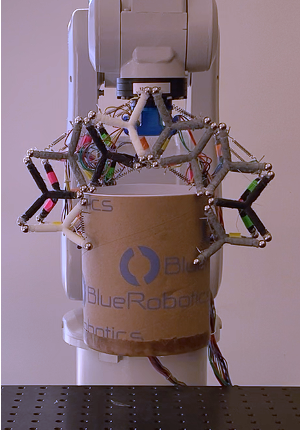}
      \caption{} 
      \label{fig:g3} 
    \end{subfigure}%
    \begin{subfigure}{.16\linewidth}
    \addtocounter{subfigure}{-1}
     \renewcommand\thesubfigure{\alph{subfigure}4}
      \centering
      \includegraphics[width=\linewidth]{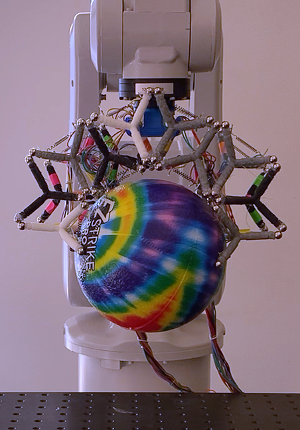}
      \caption{} 
      \label{fig:g4} 
    \end{subfigure}%
       \begin{subfigure}{.16\linewidth}
    \addtocounter{subfigure}{-1}
     \renewcommand\thesubfigure{\alph{subfigure}5}
      \centering
      \includegraphics[width=\linewidth]{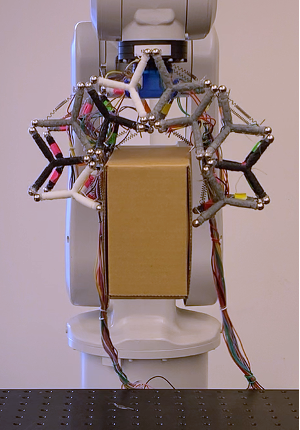}
      \caption{} 
      \label{fig:g5} 
    \end{subfigure}%
       \begin{subfigure}{.16\linewidth}
    \addtocounter{subfigure}{-1}
     \renewcommand\thesubfigure{\alph{subfigure}6}
      \centering
      \includegraphics[width=\linewidth]{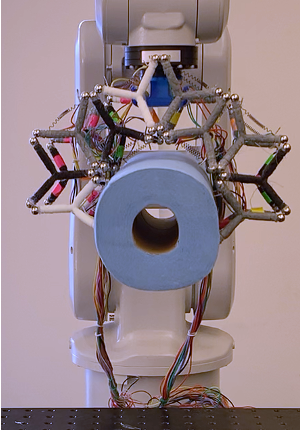}
      \caption{} 
      \label{fig:g6} 
    \end{subfigure}
    \addtocounter{subfigure}{-1}
      \caption{\hlsecondversion{Module-base gripper}}
      \label{fig:system_overview_gripper}
    \end{subfigure}
    \caption{System overview}
     \vspace{-1\baselineskip}
    \label{fig:system_overview}
\end{figure}

% Design overview
\secondversion{In this paper, we introduced soft lattice modules which embody the benefits of both soft and modular robots.}The modules in this paper are composed of 3D printed skeleton using a flexible Thermoplastic Polyurethane (TPU) material, shaped like a pair of stacked tetrahedrons (Fig.~\ref{fig:system_overview_singlemodule}), and SMA coils which are tethered to an Arduino located externally. SMAs are selected as actuators of this modular system due to their high force-to-mass ratio and high work density. The modules attach together via specially arranged magnets located at the end of the module arms.

% Our system is assembled from simple robotic modules that are 3D printed using a flexible Thermoplastic Polyurethane (TPU) material and shaped like a pair of stacked tetrahedrons (Fig.~\ref{fig:system_overview_singlemodule}). The robot modules are actuated using SMA coils which are tethered to an Arduino and control system located externally.

% \hl{todo: descripe peristaltic manipulation. Can mention contact juggling (Michael Moschen is an American juggler... ).}

% How does it work? What's in the paper?

% a tether-less system is a goal for future work \hl{should move this sentence to the future work section}. 

We designed two methods of locomotion for the individual module based off the principle of shifting the center of mass\secondversion{in order}to change the relative friction of the bottom three feet as they slide along a surface. Once individual modules approach each other, they engage in an attachment protocol that helps ensure all four contact points are connected (Fig.~\ref{fig:system_overview_attachment}). We also developed a walking gait for locomotion of three-module units that is used to assemble a larger lattice (Fig.~\ref{fig:system_overview_lattice_assembly}). Once the lattice is assembled, a peristaltic control algorithm is used to roll one or more balls along the lattice surface.  (Fig.~\ref{fig:system_overview_manipulation}). 

Peristaltic motion has been used in the robotics field as a method of locomotion~\cite{Omori2009PeristalticEarthworms, Seok2010PeristalticLocomotion}, but little information is available about the use of peristalsis (whole-body flexing) by robots for object manipulation. We explore such manipulation using the flexible surface in a way that resembles contact juggling performed by humans. \hlsecondversion{To further explore the capabilities of these lattice modules we also tested prehensile module-based gripper capable of grasping various objects. {(Fig.~\ref{fig:system_overview_gripper}).} }

% contributions
\secondversion{The}\hlsecondversion{Our}primary contribution\secondversion{of this work}is the instantiation of semi-soft lattice modules capable of attachment and detachment on-demand, enabling a rich application space including self-assembled structures and multi-functional robots. Taking inspiration and benefits from soft robotics, modular robotics, and tensegrity robotics, our proposed soft lattice modules locomote independently and collectively. Collections of modules achieve faster locomotion,\secondversion{than single modules)}and\secondversion{also}permit various manipulation strategies.
%The main contributions of the work are a new soft modular robot design, and locomotion and manipulation strategies for individual modules and collections of modules. 
%Collections of the modules further demonstrate the capability of traversing objects along the surface of an assembled lattice.
%manipulating multiple spheres along the surface of the object
%, and we show that collections of modules also achieve faster locomotion than single modules. We are particularly excited by the possibility of ultimately using controlled deformation of different module arrangements to achieve new types of locomotion and manipulation.  There are limitations to the work: for now, the system is tethered, and locomotion and manipulation strategies are open loop.

%. Next steps include: a tetherless system with on-board computing, power, sensing and communication; module designs that allow faster single-locomotion; hardware and techniques for self-disassembly; simulation reflecting the dynamics of the system; and closed-loop control to achieve more complex tasks. 

This paper is organized as follows. First, we discuss related work,\secondversion{Then, we describe}\hlsecondversion{and}the design\secondversion{and experimentation}of our system,\secondversion{in the order of that tasks that is must complete}starting with a description of the single module\secondversion{design and}locomotion, followed by attachment into 3-module,\secondversion{units}followed by multi-module locomotion and attachment into a lattice, and then by peristaltic manipulation\secondversion{Lastly, we draw conclusions,}\hlsecondversion{and module-base gripper.}We also discuss limitations of the current system, and propose future work.

% % TODO: move this into the single module design section.
% We referred the geometry of tetrahedral molecular(shown in figure~\ref{fig:molecular}) and designed the structure of a single modular robot to be a symmetric structure vertically to achieve both locomotion and manipulation( figure \ref{fig:single_design} ). We printed each module with rubber-like materials where each rod of the module has the properties that a spring has.
% actuator
% Given the size of the module is small and each module needs 9 actuators to achieve all the tasks, we used one-way SMA(shape memory alloy) coils to actuate the robots. The SMA's length  shortened when heated and return to its original length/shape when cooled down, where the property can be applied to our robot to change its own shape to achieve the task.
% locomotion
% The principle of locomotion we explored is to switch the center of mass to weaken the friction when the direction of friction is backward and reinforce the friction when its direction is forward for both single module locomotion and multiple module locomotion. The only difference is that given each module has 3 feet on the ground, if one foot has been lifted, it will fall down, therefore the feet for single module locomotion could not be lifted but can be lifted when 3 modules have been attached together. 

% \hl{todo: paragraph on the attachment process, steps towards forming a connected lattice. Include the ``3-legged race" multi module locomotive ability.}

% \hl{todo; paragraph on peristaltic manipulation. Mention 2d \& modeling as developmental work.}

\section{Related work}
\label{sec:related}

Our work lies at the intersection of soft robots and modular robots and relates to much other work in these fields.\secondversion{In particular, our system is a form of a self-assembling and re-configuring robot, which is in contrast to many modular robots that require manual assembly and reconfiguration.}Another field of related work is tensegrity robots, in which the robot structure is held intact through a combination of compressive and tensional elements. Although our modules and system do not rely on tensional forces when static, they behave much like  tensegrities when an SMA spring is activated.  \secondversion{We now discuss related work in each of these related categories.}

\subsection{Modular Soft robots}

Lee et al.~\cite{lee2017soft} provide a review of soft robot research, including an overview of fabrication methods (ours is 3D printed), actuation methods (ours uses SMA wire), and control techniques.\secondversion{However, we are not aware of thorough reviews of modular soft robots; While we do not include a thorough review of modular soft robots, we note a few recent examples.}Several modular soft robots are composed of repeating sections, often arranged in caterpillar-inspired shapes and commanded with traveling peristaltic waves. For example, the Zou catepillar robot~\cite{zou2018reconfigurable} used modular segments with pneumatic feet connected in series and the Wormbot~\cite{nemitz2016using} featured custom audio speaker assemblies embedded in sections. Both locomoted using peristaltic control strategies.

The use of magnetic attachment is also fairly common among modular \hlsecondversion{rigid}and soft robots. \hlsecondversion{Tosun et al. designed a connector embedded in planar face for rigid modular robots using electro-permanent magnets where power is needed for changing magnets state{~\cite{7759033}}.}The Kwok self-assembling finger robot~\cite{kwok2014magnetic} demonstrated magnetic self assembly of robotic fingers to form hexapods and grippers manually and remotely. The CBalls modules~\cite{CBalls_Chen_2019} are soft robots composed of pairs or trios of inflatable spherical bodies magnetically attached together and driven through a rolling motion between\secondversion{the}bodies.

% \hlsecondversion{Other modular soft robot include Jellocube, a soft robot that can jump individually~{\cite{Li2019JelloCubeAC}}. Usevitch et al.~{\cite{isoperi}} designed a modular shape soft robot that is untethered and isoperimetric, which allows for shape-changing, punctuated rolling and manipulation.}
There are other soft robots described as modular, but which use the word differently than we are using in this work. The ``Click-e-bricks'' components~\cite{morin2014using} and The Limpet II \cite{SayedEtAlLimpetII} are both\secondversion{The ``Click-e-bricks'' components~\cite{morin2014using} are reminiscent of plastic brick toys which are constructed and then glued together with additional elastomer. The Limpet II \cite{SayedEtAlLimpetII} is}described as modular, but this term is used to describe interchangeable components, such as sensors or actuators, and not that the individual robots work together as a collective. The Omniskins Prior ``robotic skins'' (or omniskins)~\cite{booth2018omniskins} are reconfigurable components that afford new shapes rather than collective robot swarms that can link together.

% Another major distinction between this work and prior is that, with the exception of Kwok finger robots, none of these robots demonstrated self-assembly. The Kwok finger robots and the Omniskins are used to demonstrate manipulation, but as grippers on manipulators, not as 2-D surfaces.

Another major distinction between this work and prior is that,\secondversion{with the exception of}\hlsecondversion{except}Kwok finger robots, none of these robots demonstrated self-assembly, \hlsecondversion{including self-locomotion and attachment.}\secondversion{The Kwok finger robots and the Omniskins are used to demonstrate manipulation, but as grippers on manipulators, not as 2-D surfaces.}\hlsecondversion{Jellocube is a soft robot that can jump individually~{\cite{Li2019JelloCubeAC}}. Usevitch et al.~{\cite{isoperi}} designed a modular shape soft robot allowing for shape-changing, punctuated rolling and manipulation, but none of them show attachment functionality.}

\hlfinalversion{Although rolling or jumping locomotion might be faster}\cite{Li2019JelloCubeAC,isoperi,CBalls_Chen_2019},
\hlfinalversion{in this paper, we focus on walking and sliding motions as they provide the needed precision for docking, as well as multi-module locomotion.}

\subsection{Self-reconfiguring robots}

Yim et al.~\cite{Yim2002ModularRobots} and more recently Parker et al.~\cite{Parker2016} provide reviews of re-configurable and self-assembling robots\hlsecondversion{, and points out important challenges of self-configuring robot systems{~\cite{4141032}}.}These robots often assemble into a chain or lattice arrangement. \hlsecondversion{Yim et al. first presented locomotion gaits table of rigid modular re-configurable robots{~\cite{yim}}.}
% \hlsecondversion{Yim et al.{~\cite{4141032}} also points out some important challenges of self-configuring robot systems and proposes some locomotion gaits of rigid modular re-configurable robots{~\cite{yim}}}. 

% \hlsecondversion{Liu et al. presented a reconfiguration planning algorithm for SMORES modular robots to behave self-reconfiguration in a distributed manner{~\cite{8769941}}.}
\secondversion{Wei et al. describe Sambot, a rigid body chain-based reconfigurable robot where both the modular components and assembled chain are capable of locomotion.} 
\hlsecondversion{SMORES{~\cite{8769941}} and}Sambot~\cite{wei2010Sambot} modules are un-tethered and fully autonomous reconfigurable robot (our system is tethered), but are rigid-bodied and \hlsecondversion{requires additional part for manipulation capability.}\secondversion{have not demonstrated manipulation capability.Suzuki et al. describe}ShapeBots~\cite{ShapeBotsSuzukiEtAl}\secondversion{, which }are shape-changing robots capable of working together to perform object manipulation,\secondversion{data visualization}\hlsecondversion{but cannot fully deform because the robots are rigid.}
\secondversion{Although shape-changing, ShapeBots are not soft robots, as their core structure is a rigid boxand only a portion of the robot extends. Additionally, ShapeBots do not directly connect together.}

\hlsecondversion{Motion planning is a significant challenge for a large amount of modular robot and reconfigurable robot.}Kilobots \cite{rubenstein2012kilobot} collectively make up a 1024-robot system capable of self-assembly.\secondversion{, and intended to serve as a testbed for distributed AI systems. Like our system, the kilobot swarm is composed of many individual robots with limited capabilities. Our system differs from the kilobot swarm in the soft robot body, SMA-based actuation mechanism, and magnetic inter-robot attachment.} \secondversion{The individual robots in our system work together in a mechanical sense to achieve manipulation.}\hlsecondversion{Liu and Yim proposed a motion planning algorithm for self-reconfiguration ability of variable topology truss{~\cite{8967640}}.}

\subsection{Tensegrity Robots}

The module design draws some inspiration from tensegrity structures~\cite{skelton2001introduction}.\secondversion{Tensegrity structures are lightweight, durable, energy efficient and compliant {\cite{skelton2001introduction}}, and therefore have unique advantages and applications over other traditional types of autonomous robots. As tensegrity robots are difficult to control due to their complex structure, one of the greatest challenges in the emerging field of tensegrity research is finding effective methods for tensegrity automation.}There is much prior research surrounding tensegrity robots and their applications, following on seminal work at NASA on the SUPERball in 2014. Due to its light weight, it is especially well-adapted to maneuvering over hazardous terrain~\cite{bruce2014superball,sabelhaus2015system,bruce2014design}.

\hlsecondversion{Soft tensegrity robots have also been explored for different purposes. Rieffel and Mourel accomplished fast locomotion using soft tensegrity robot actuated with 3 motors{~\cite{Rieffel}}. Lee et al. fabricated soft tensegrity robots with 3D printing materials to avoid additional pose-assembly process for single robot{~\cite{hajunlee}}. Zappetti et al. proposed a soft modular robot inspired by cytoskeleton of living cell based on tensegrity{~\cite{Zappetti2017BioinspiredTS}}.}

Soft biologically-inspired tensegrity robots, built with rods, tension springs and  motorized cables, lend themselves to many different modes of motion \cite{he2020adaptable,lessard2016bio,boxerbaum2012continuous}. These tensegrity structures are known as ``bio-tensegrity robots,'' and take inspiration from a variety of animals. \secondversion{For example, some robot designs have been based on a form of ``earthworm-like'' peristaltic motion to achieve locomotion through tightly constrained spaces \cite{boxerbaum2012continuous}. The worm robot is segmented into small tensegrity modules which are permanently attached to each other (thus it is not a modular tensegrity system).Then, this peristaltic motion is informed by a rolling-wave algorithm that continuously passes a periodic motion across the segments of the body of the worm robot by contracting the tensile connectors that occur between segments.}\hlsecondversion{Mirletz1 et al. used NASA Tensegrity Robotics Toolkit (NTRT) to simulate morphology and control of modular spine-like tensegrity robots{~\cite{spine-like}}, where they tested on Octahedral and Tetrahedal complexes, which looks similar to our module, but their module cannot locomote individually or do self-assembly. Moreover, the connection between their modules are strings, and ours are magnets, which means our single module is more flexible and separable.}

\secondversion{Simple microscopic robots have also been developed that utilize inchworm-like friction-based locomotion~\cite{donald2008self}. This methodology could be used to inform the motion of microscopic bio-tensegrities, and be particularly useful in developing locomotion protocols for individual tiny tensegrity modules.}

There is also much prior work on the coordination of non-tensegrity multi-robot systems and\secondversion{the use of}\hlsecondversion{using}these systems for object manipulation. There are many advantages to multi-robot systems~\cite{arai2002advances}, and they are particularly well suited for the transportation of large objects in hazardous environments. There are a variety of approaches to coordinate the motions of robots in these systems. For example, cluster space control~\cite{mas2012object} relies on a pilot to use a joystick and manually coordinate four robots. These robots can\secondversion{then}be used together for object transportation. Decentralized control methods have also been studied that allow robots to identify and approach a target object and move around to transport it together~\cite{song2002potential}. The majority of such approaches rely on the robots themselves to travel distances and move around the surface, but there is less existing work on static robots\secondversion{that are able}to pass objects between them.

\section{\hlsecondversion{Module Design}}

\subsection{\hlsecondversion{Hardware Design and Assembly}}
An individual module is composed of a skeleton, SMAs and spherical magnets. The shape of the skeleton is a pair of symmetric stacked tetrahedrons, inspired by the geometry of a tetrahedral molecule (Fig.~\ref{fig:system_overview_singlemodule}). \hlsecondversion{Each tetrahedron is composed of four equal-length bars. The angle between each two neighboring bars is $\cos^{-1}(-1/3) \thickapprox 109.5^{\circ}$, which is the same as the bond angle of the tetrahedron molecule. The diameter of each bar is smaller near the connection point for better flexibility, with grooves for mounting SMAs. The skeleton is printed with flexible TPU material with an infill density of 90\% to achieve a stronger adversarial bending force that helps restore a module's original shape after SMA contraction.}\secondversion{We 3D printed each module with flexible TPU plastic, allowing the module to bend when a force is applied.}

Given the small size of each module and the need for nine actuators in order to achieve our target locomotive gaits and manipulation capabilities, we used one-way SMA (shape memory alloy) coils (coil diameter, $3.45~mm$; wire diameter, $0.51~mm$, Dynalloy) to actuate the robots. \hlsecondversion{The end of each spring SMAs is crimped with fishing wire and electric wire in a ferrule and mounted to the grooves in the skeleton with the fishing wire. The detwinned Martensite rest length of each SMA is longer than the distance between two grooves to make sure that they can be extended when the skeleton deforms because of SMA actuation.}The Austenitic (actuated) rest length of the actuators is $1.52~cm$/$1.42~cm$/$0.91~cm$ (vertical/diagonal/horizontal). When installed, the detwinned Martensite rest length is set to be $8.0~cm$/$7.6~cm$/$4.4~cm$, though this exact length changes with actuation cycles due to variance in antagonistic forces applied by other SMA actuators and the elastic energy stored in the deformed soft skeleton.

Spherical magnets are affixed at the end of each tetrahedral limb to enable the modules to connect together. We oriented the polarity of the magnets in a circle as described in \cite{Zwier2017MagneticSC}, where each magnet in the loop is aligned head-to-tail and the magnetic field inside the loop is either clockwise or counterclockwise. Therefore, once all the magnetic balls are glued on the modules with the same magnetic field direction, any modules can then be attached together from all directions. 

\begin{figure}[h]
\centering
\begin{subfigure}{.3\linewidth}
  \centering
  \includegraphics[width=\linewidth]{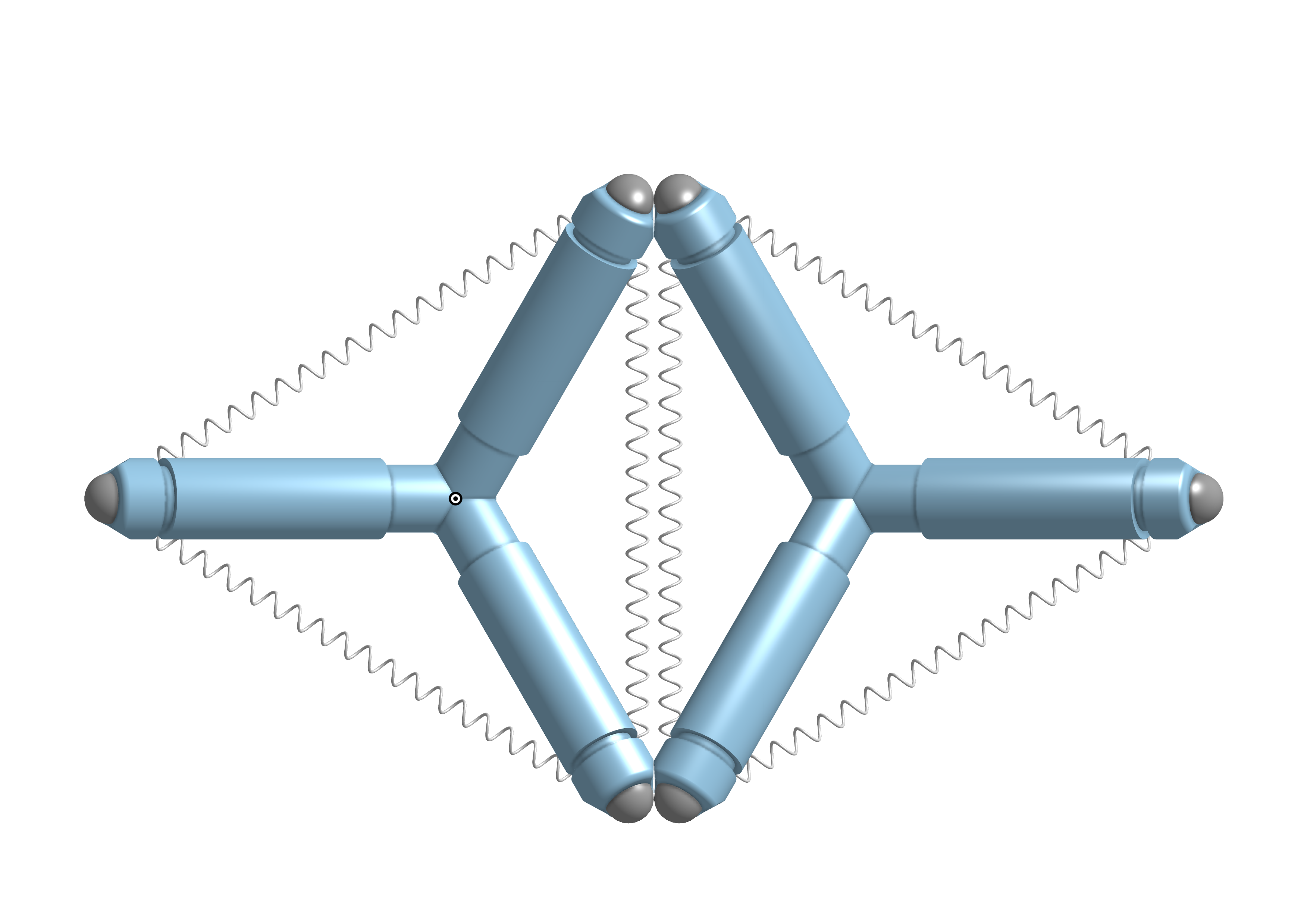}
  \caption{}
%   \caption{Four feet connected}
  \label{fig:two_point_connected}
\end{subfigure}%
\begin{subfigure}{.3\linewidth}
  \centering
  \includegraphics[width=\linewidth]{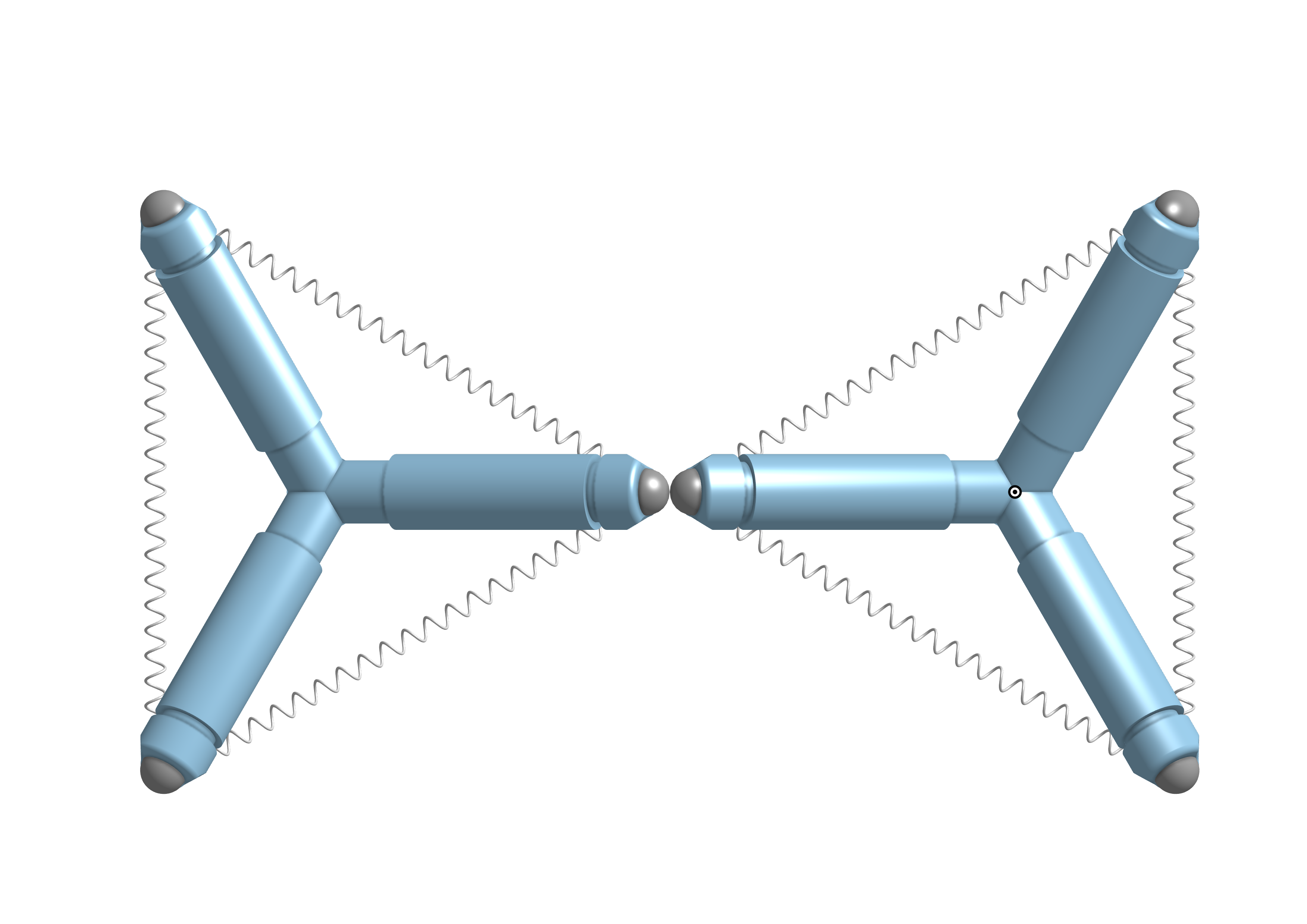}
  \caption{}
%   \caption{Two feet connected}
  \label{fig:one_point_connected}
\end{subfigure}
\caption{Top view of 2 pairs and 1 pair magnetic connectors}
\vspace{-3mm}
\label{fig:attachment_evaluation}
\end{figure}

\subsection{\hlsecondversion{Choice of spherical magnets}}
\hlsecondversion{The pull force of the spherical magnet has a strongly influences on the attachment property. We tested four sizes of spherical magnets: 3/16", 1/4", 5/16", and 3/8". Six experiments were done to evaluate their attachment properties including the maximum distance that two modules can be attached by magnetic force only and the minimum length that SMAs are allowed to be contracted to avoid detachment of magnetic connectors. The maximum distances were tested in two cases: four feet facing each other (8 magnets) and two feet facing each other (4 magnets). The minimum contraction length were tested in four cases: two feet connected (Fig.{~\ref{fig:two_point_connected}}) and four feet connected (Fig.{~\ref{fig:one_point_connected}}) while only SMAs in one module contracted and SMAs in two modules contracted. Experimental results are shown in Table.{~\ref{table:attachment_evaluation_result}}, where $1.52~cm$ is the Austenitic rest length of the vertical SMA. Results indicate that the magnetic connection does not break even with maximum SMA actuation. Based on the results, we chose the 1/4" diameter for locomotion and peristaltic manipulation because of its low cost, and chose the 5/16" magnets for gripper experiments that required larger deformation while keeping modules connected.}

\begin{table*}[h]
\centering
\caption{Attachment and Detachment Evaluation}
\begin{tabular}{ |c|c|c|c|c|c|c|c| } 
 \hline
  \textbf{\thead{Diameter\\ (inch)}}
  & \textbf{\thead{Pull \\Force (lb)}}
  & \textbf{\thead{Max attach \\distance for \\ 8-magnets (cm)}}
  & \textbf{\thead{Max attach \\distance for \\ 4-magnets (cm)}}
  &
  \textbf{\thead{Min length\\ for 2 SMAs \\ in Fig.~\ref{fig:one_point_connected} (cm)}}
  & \textbf{\thead{Min length\\  for 1 SMA\\ in Fig.~\ref{fig:one_point_connected} (cm)}}
  & \textbf{\thead{Min length\\  for 4 SMAs \\in Fig.~\ref{fig:two_point_connected} (cm)}} 
   & \textbf{\thead{Min length\\ for 2 SMAs\\ in Fig.~\ref{fig:two_point_connected} (cm)}} 
    \\ 
  \hline
  3/16  & 1.04 & 1.5 & 1 & 1.52 & 4 & 5.5 & 5.5\\ 
  \hline
  1/4  & 1.86 & 2 & 1.5 & 1.52 & 1.52 & 4 & 4\\ 
  \hline
  5/16  & 2.52 & 3 & 2 & 1.52 & 1.52 & 2 & 2\\ 
  \hline
  3/8  & 3.69 & 5 & 3 & 1.52 & 1.52 & 1.52 & 1.52\\ 
 \hline
\end{tabular}
\label{table:attachment_evaluation_result}
\end{table*}

\begin{figure*}[b]
\vspace*{-3mm}
\centering
  \includegraphics[width=\linewidth]{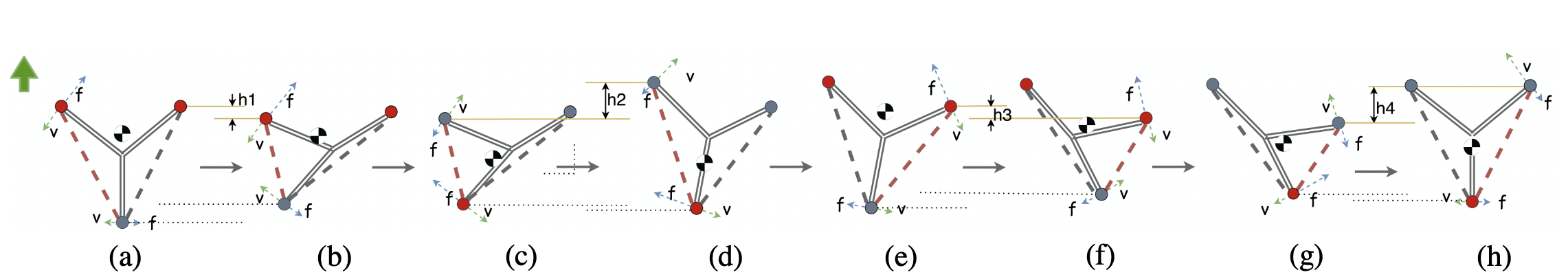}
\caption{Physical analysis of {\em Shuffling} pattern locomotion (top-down view of bottom legs)}
\label{fig:shuffling_physical}
\end{figure*}

\begin{figure}[h]
\centering
 \vspace{-1.5\baselineskip}
  \includegraphics[width=\linewidth]{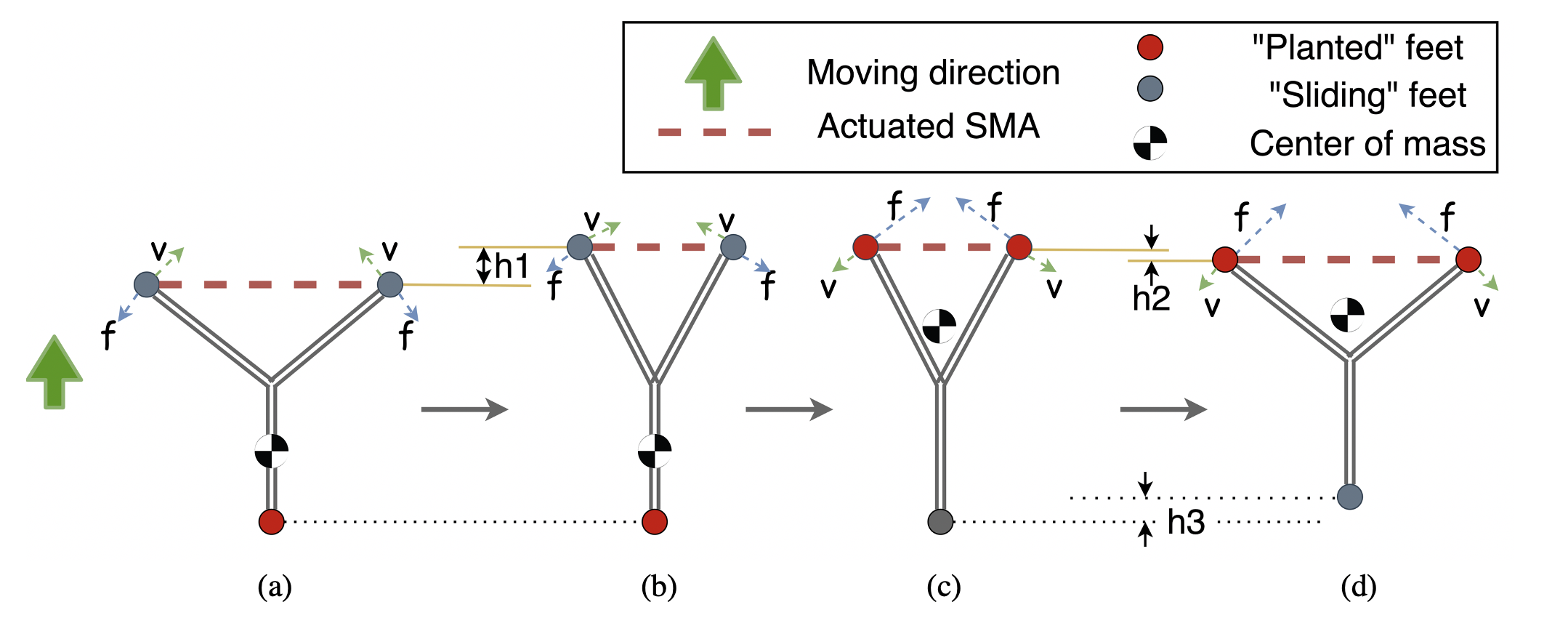}
\caption{ \centering Physical analysis of {\em Grab-and-pull} pattern locomotion \newline (top-down view of bottom legs)}
 \vspace{-1.5\baselineskip}
\label{fig:grab-and-pull_physical}
\end{figure}

% % 0.184
% \begin{figure*}[htb]
% \begin{center}
% \begin{subfigure}{0.1656\textwidth}
%   \includegraphics[width=\linewidth]{images/single1.png}
%   \caption{Initial state} \label{fig:s1}
% \end{subfigure}%
% \begin{subfigure}{0.1215\textwidth}
%   \includegraphics[width=\linewidth]{images/single2.png}
%   \caption{Front down} \label{fig:s3}
% \end{subfigure}%
% \begin{subfigure}{0.1215\textwidth}
%   \includegraphics[width=\linewidth]{images/single3.png}
%   \caption{Contract left} \label{fig:s3}
% \end{subfigure}%
% \begin{subfigure}{0.1215\textwidth}
%   \includegraphics[width=\linewidth]{images/single4.png}
%   \caption{Back down} \label{fig:s4}
% \end{subfigure}%
% \begin{subfigure}{0.1215\textwidth}
%   \includegraphics[width=\linewidth]{images/single5.png}
%   \caption{Release left} \label{fig:s5}
% \end{subfigure}%
% \begin{subfigure}{0.1215\textwidth}
%   \includegraphics[width=\linewidth]{images/single6.png}
%   \caption{Contract front} \label{fig:s6}
% \end{subfigure}
% \caption{Locomotion example of combining {\em Grab-and-pull} and {\em shuffling} patterns}
% \label{fig:3_leg_race}
% \end{center}
% \end{figure*}

\section{Single Module Locomotion \& Attachment}
\label{sec:single-loco}

% \begin{figure}[h]
% \centering
% \begin{subfigure}{.3\linewidth}
%   \centering
%   \includegraphics[width=\linewidth]{images/mag1.jpeg}
%   \caption{Clockwise}
%   \label{fig:mag1}
% \end{subfigure}
% \hspace*{0.1\linewidth}
% \begin{subfigure}{.3\linewidth}
%   \centering
%   \includegraphics[width=\linewidth]{images/mag2.jpeg}
%   \caption{Counterclockwise}
%   \label{fig:mag2}
% \end{subfigure}
% \caption{Hexagonal Loop of Magnets in Two Directions}
% \label{fig:mag_balls}
% \end{figure}

\secondversion{An individual module is composed of a skeleton and SMAs, where we designed the structure of the skeleton as a pair of stacked tetrahedrons inspired by the geometry of a tetrahedral molecule.
The modules' skeletons are vertically symmetric in order to achieve both locomotion and manipulation (Fig.{~\ref{fig:system_overview_singlemodule}}). We 3D printed each module with flexible TPU plastic, allowing the module to bend when a force is applied. Given the small size of each module and the need for nine actuators in order to achieve our target locomotive gaits and manipulation capabilities, we used one-way SMA (shape memory alloy) coils (coil diameter, $3.45~mm$; wire diameter, $0.51~mm$, Dynalloy) to actuate the robots. The Austenitic (actuated) rest length of the actuators is $1.52~cm$/$1.42~cm$/$0.91~cm$ (vertical/diagonal/horizontal). When installed, the detwinned Martensite rest length is set to be $8.0~cm$/$7.6~cm$/$4.4~cm$, though this exact length changes with actuation cycles due to variance in antagonistic forces applied by other SMA actuators and the elastic energy stored in the deformed soft skeleton.}

\secondversion{Spherical magnets are affixed at the end of each tetrahedral limb to enable the modules to connect together. We oriented the polarity of the magnets in a circle as described in {\cite{Zwier2017MagneticSC}}, where each magnet in the loop is aligned head to tail and the magnetic field inside the loop is either clockwise or counterclockwise. Therefore, once all the magnetic balls are glued on the modules with the same magnetic field direction, any modules can then be attached together from all directions.}

% The SMA's length  shortens when heated via an electric current and returns to its original length/shape when cooled down below austenite finish temperature (~60). 
%The SMA's length is stretched properly when attaching to the soft skeleton as the initial state, shortens to original length/shape when heated via an electric current, and returns to initial state under the antagonistic force provided by the elastic energy stored in the deformed soft skeleton.

\begin{table*}[h]
\centering
\caption{Running Time Comparison Between Different Patterns (Moving $2~cm$)}
\begin{tabular}{ |c|c|c|c|c|c|c|c| } 
 \hline
  \textbf{Pattern} & \textbf{{\em Grab-and-pull}} & \textbf{{\em Shuffling}} & \textbf{{\em Combined}} &  \textbf{\thead{ {\em Combined} \\ + Vibration}} & \textbf{\thead{ {\em Combined} \\ + Vibration \\+ Cooling Fan}} &  
  \textbf{\thead{ {\em Combined}  \\ + Vibration \\ with Thinner SMAs}} & 
  \textbf{\thead{3-module \\Locomotion}} \\
 \hline
  \thead{Time\\ (mins)}  & $8.30~$ & $5.02~$ & $3.15~$ & $2.76~$ & $2.58~$ & $2.51~$ & $2.35~$\\ 
 \hline
\end{tabular}
\label{table:sm-speed-results}
\end{table*}

\subsection{Analysis of Locomotive Gaits}

We have chosen two gaits (and a combination of them)\secondversion{to explore}for single-module locomotion. The purpose of locomotion for single module is to move towards other modules to assemble into a 3-module unit for locomotion, where 3-module units are utilized to assemble into a larger structure for manipulation. While there are gaits that roll modules end-over-end to achieve fast locomotion, the need for precise control during attachment motivated us to instead explore gaits that cause the robot to slide along the floor in an upright configuration.

Both gaits operate on the same principle: the module is controlled in order to move the center of mass to reduce the friction at contacts that are sliding forwards, while increasing friction at contacts that are planted. % and pushing the robot forwards. 

\subsubsection{Single-module Gait \#1: Grab-and-pull}

In this {\em grab-and-pull} gait (akin to breaststroke), the first step shifts the center of mass backwards to plant the back leg while contracting the front SMA to slide the front legs forward (Fig.~\ref{fig:grab-and-pull_physical}a). 
% alone without actuating other SMAs, as the red dotted line shows in Figure \ref{fig:grab-and-pull_physical}(a).  
% When this SMA wire is contracted, the friction force is is directed backwards with respect to the direction of travel, and when released, the friction force is forward. Since friction force depends on mass and roughness of the materials and these two are constant, the friction forces during contraction and release are the same, an the robot will not move during a cycle.
% To move the robot forward, our strategy is to reduce the backwards friction force while the front SMA contracts, by shifting the mass to be over the back foot, marked in red in the figure. 
Similarly, during release of the front SMA, the robot rocks so that the center of mass lies over the front two feet (Fig.~\ref{fig:grab-and-pull_physical}b). Therefore, after switching the center of mass from back to front, the distance moving forward\secondversion{will be equal to}$h_3$ (Fig.~\ref{fig:grab-and-pull_physical}b)\secondversion{, which}is equal to $h_1 - h_2$, where $h_1$ and $h_2$ are the same when the center of mass is not switched and kept in the center. 

% \hl{Should we mention that the length of a force arrow represents the magnitude of the force?}
\subsubsection{Single-module Gait \#2: Shuffling}

\secondversion{The other pattern is named {\em shuffling}, where}\hlsecondversion{In this pattern,}the robot walks by dragging left and right feet alternately without lifting the feet from the ground. The physical analysis behind it is similar to the {\em grab-and-pull} pattern. Fig.~\ref{fig:shuffling_physical} shows the whole process, where the center of mass is switched to front during contraction of left SMA and switched backward during recovery process to intensify the impact of forward friction and 
weaken the impact of backward friction. Then the right SMA works in the same way symmetrically. 

Importantly, if the contraction time of left and right SMAs are different, the robot can rotate to the direction that is less contracted, which means with this pattern, the upright robot can move to arbitrary locations and orientations in the plane.

\subsubsection{Single-module Gait \#3: Combined }

\begin{figure}[h]
\vspace*{-3mm}
\centering
  \includegraphics[width=0.6\linewidth]{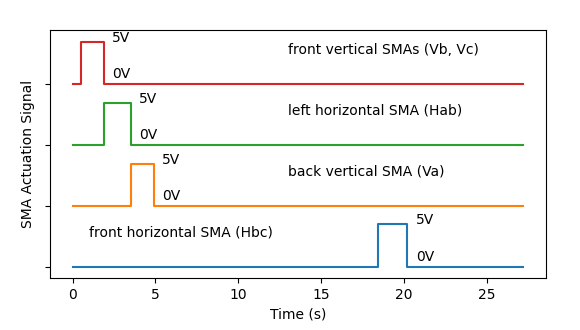}
\caption{Control signal for single module locomotion}
 \vspace{-0.5\baselineskip}
\label{fig:single_module_control_signal}
\end{figure}

\begin{figure}[htb]
\centering
\begin{subfigure}{0.2116\linewidth}
   \includegraphics[width=\linewidth]{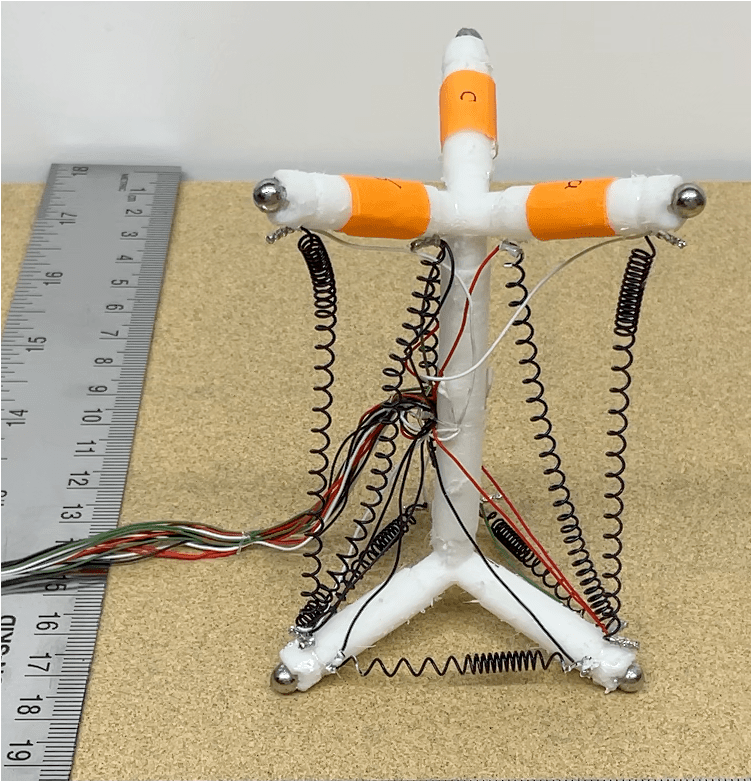}
   \caption{} \label{fig:s1}
\end{subfigure}%
\begin{subfigure}{0.15525\linewidth}
   \includegraphics[width=\linewidth]{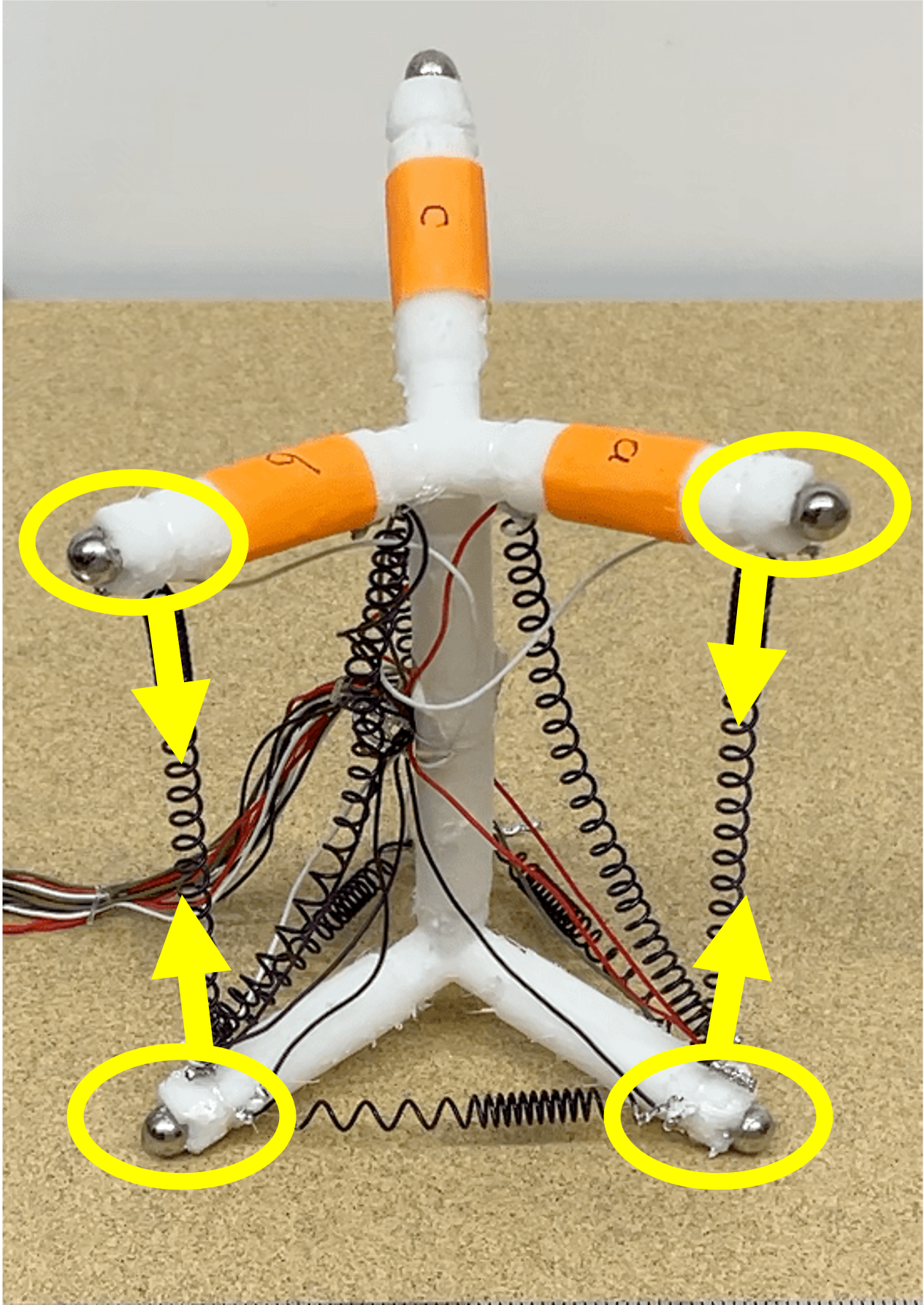}
   \caption{} \label{fig:s3}
\end{subfigure}%
\begin{subfigure}{0.15525\linewidth}
   \includegraphics[width=\linewidth]{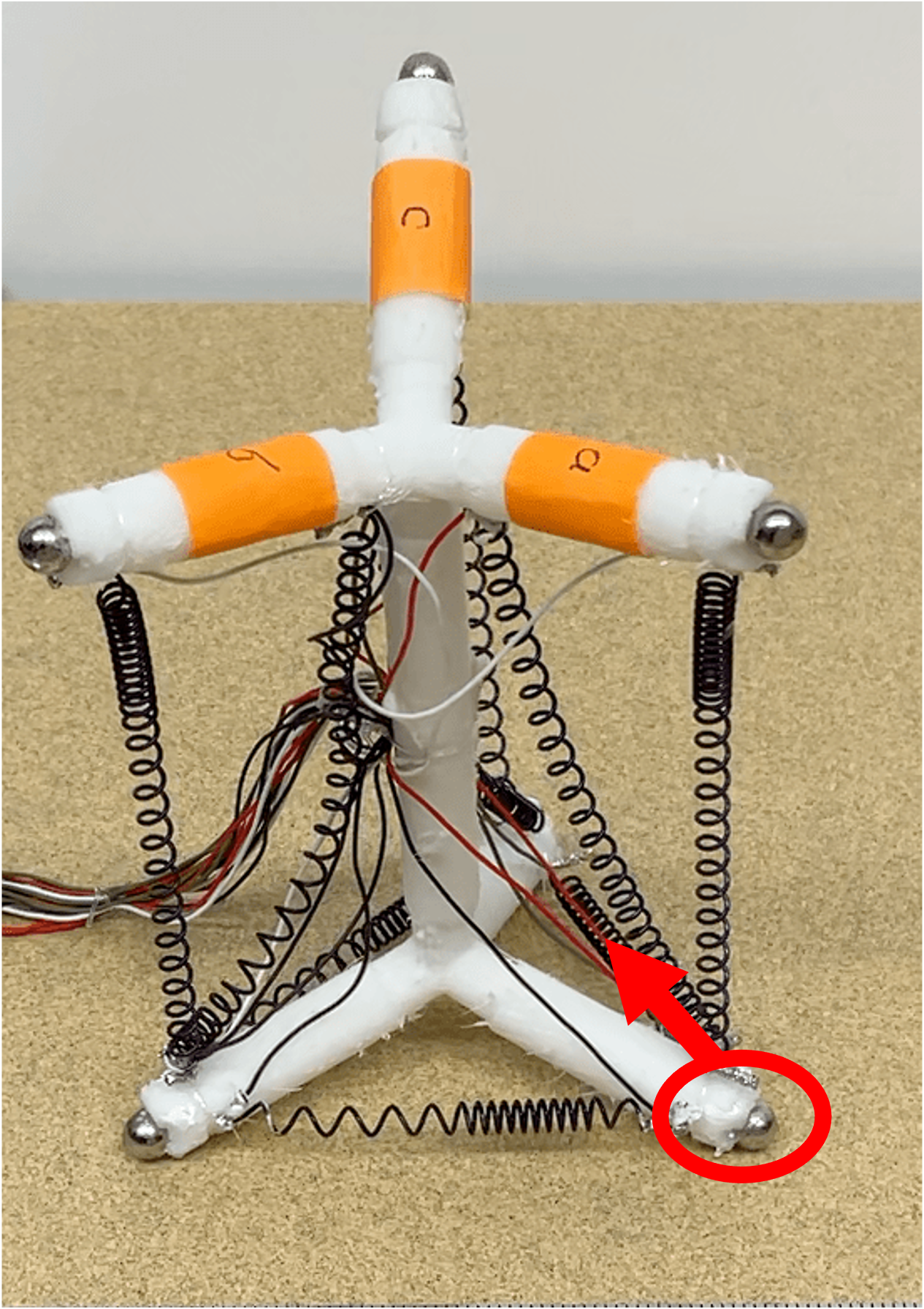}
   \caption{} \label{fig:s3}
\end{subfigure}%
\begin{subfigure}{0.15525\linewidth}
   \includegraphics[width=\linewidth]{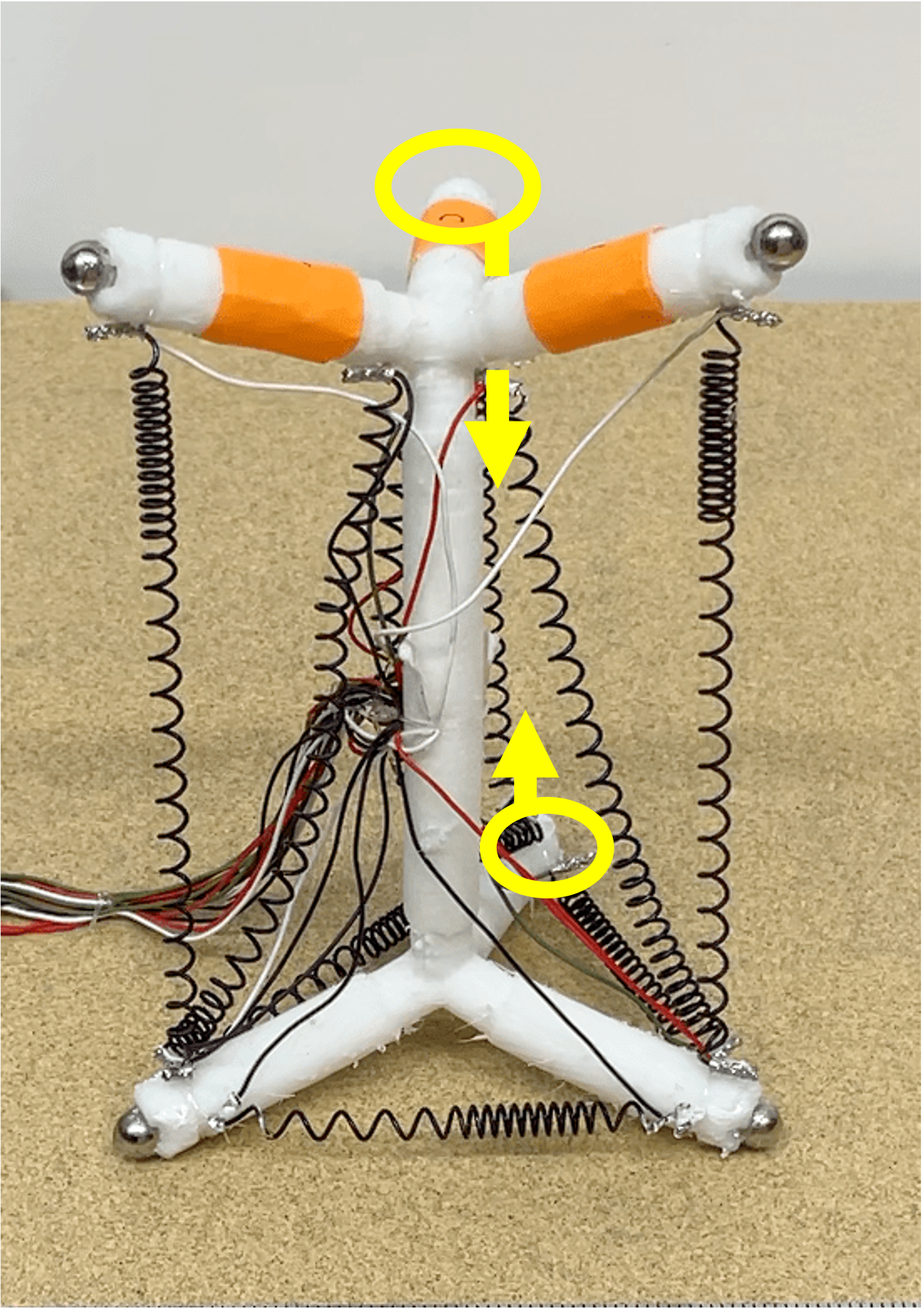}
   \caption{} \label{fig:s4}
\end{subfigure}%
\begin{subfigure}{0.15525\linewidth}
   \includegraphics[width=\linewidth]{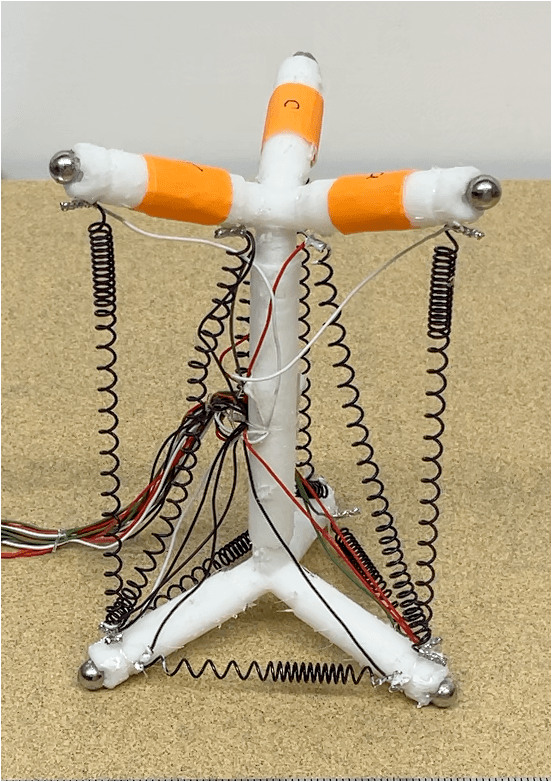}
   \caption{} \label{fig:s5}
\end{subfigure}%
\begin{subfigure}{0.15525\linewidth}
   \includegraphics[width=\linewidth]{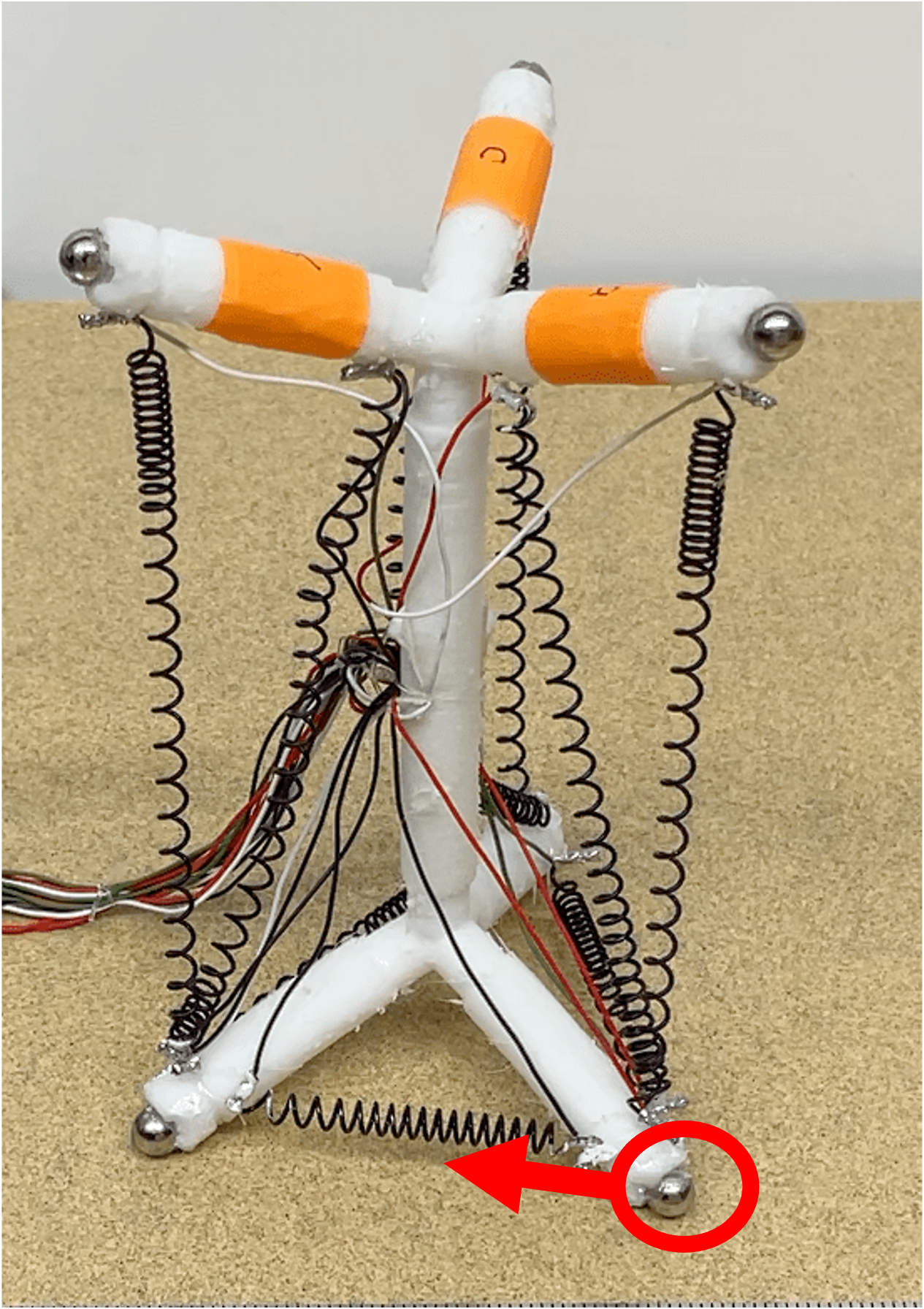}
   \caption{} \label{fig:s6}
\end{subfigure}
\caption{\secondversion{Locomotion example of}Combining {\em Grab-and-pull} and {\em shuffling} patterns}
\label{fig:3_leg_race}
\vspace{-1\baselineskip}
\end{figure}

We used the property that motion of the center of mass is opposite in the two patters to create a combined pattern with faster locomotion. Specifically, the combined gait is ``left \textrightarrow front \textrightarrow right \textrightarrow front". Fig.~\ref{fig:3_leg_race} shows an example where the center of mass switches from front to back when left foot is contracted \& released and then the front bottom SMA is contracted -- the corresponding control signal is shown in Fig.~\ref{fig:single_module_control_signal} (SMAs are labeled in Fig.~\ref{fig:system_overview_singlemodule}; The labels Vx, Dx, Hxx represent vertical, diagonal and horizontal SMAs respectively, where x/xx is determined by the limb label). We ultimately selected this combined gait for the single-module locomotion and attachment task due to its speed advantage.

\subsection{Experiments}

\subsubsection{Locomotion Performance}
To control the SMAs, we used\secondversion{an}Arduino and\secondversion{a}$12~V$ DC power supply. 
120 grit sandpaper supplies a coarse\hlsecondversion{ locomotion}surface\secondversion{for locomotion}.
To reduce drag, we used light, flexible silicone insulated stranded wires for the tether.

We conducted some experiments to validate the effectiveness of the two locomotion patterns. The {\em grab-and-pull} gait took $8.3~min$ to travel $2~cm$, or \hlsecondversion{$0.00095$}body lengths per second ($BL/s$), where $1.38~min$ was used to actuate and the rest was used to cool the SMA actuators. The {\em shuffling} pattern took $5.02~min$ to travel $2~cm$, or \hlsecondversion{$0.00157~BL/s$,}where $0.84~min$ was used to actuate. 
For the combined gait, the time cost of moving forward $2~cm$ is $3.15~min$ (median of 10 trials), or \hlsecondversion{$0.00252~BL/s$,}where the actuation time is $0.75~min$.

\subsubsection{Strategies To Increase the Speed}
Given the slow speeds observed with the gaits mentioned above, we experimented with several methods of expediting the locomotion (Table{~\ref{table:sm-speed-results}}).
% \hlsecondversion{Results are shown in Table{~\ref{table:sm-speed-results}}.}\secondversion{We did not use these techniques for our task demonstrations, but we mention them here to illustrate opportunities for speed improvements.}

\emph{Vibration}: Since legs are unable to be lifted off the ground,\secondversion{in our three-legged modules}backward friction between the sliding legs and the ground inhibits movement speed. To reduce this friction, \secondversion{we tested a vibration technique where}we alternately actuate and relax the back vertical SMAs on a $50~ms$~:~$1000~ms$ (actuation~:~cooling) interval.\secondversion{The time cost (moving $2~cm$) is $2.76~min$ using vibration on {\em combined} gait.}

\emph{Cooling Fan}: \secondversion{Our lab setting has minimal airflow, so}We tested putting a fan opposite to the direction of motion to simulate locomotion in outdoor conditions.\secondversion{The time cost (moving $2~cm$) of using {\em combined} gait, vibration and cooling fan is $2.58~min$.}

\emph{Multiple Thinner SMAs}: Approximately $5/6$ of the locomotion time is spent waiting for the SMAs to cool and expand. We tested replacing single SMAs with multiple thinner SMAs to increase the surface area and speed cooling, similar to the ``stacked-ribbon" design described in~\cite{buckner2021design}.\secondversion{The time cost (moving $2~cm$) of using {\em combined} gait, vibration with multiple thinner SMAs is $2.51~min$.}
% 0.184

\subsection{Attachment Between Single Modules}

\begin{figure}[h]
\centering
\begin{subfigure}{.33\linewidth}
  \centering
  \includegraphics[width=\linewidth]{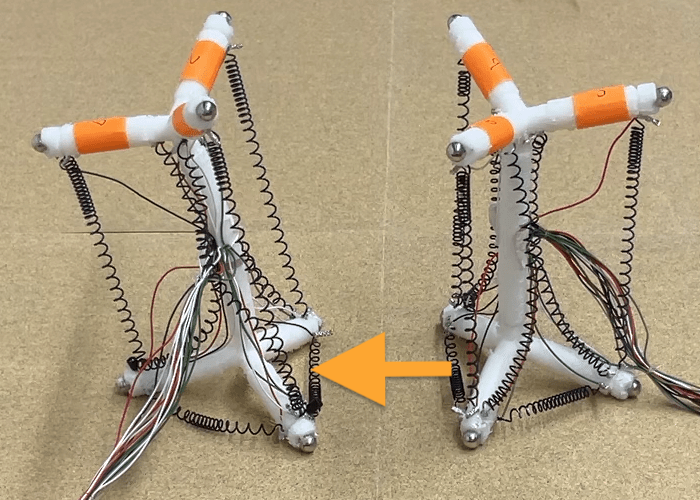}
  \caption{Move forward}
  \label{fig:a1}
\end{subfigure}%
\begin{subfigure}{.33\linewidth}
  \centering
  \includegraphics[width= \linewidth]{images/attach2-new-min.png}
  \caption{Attach bottom}
  \label{fig:a2}
\end{subfigure}%
\begin{subfigure}{.33\linewidth}
  \centering
  \includegraphics[width= \linewidth]{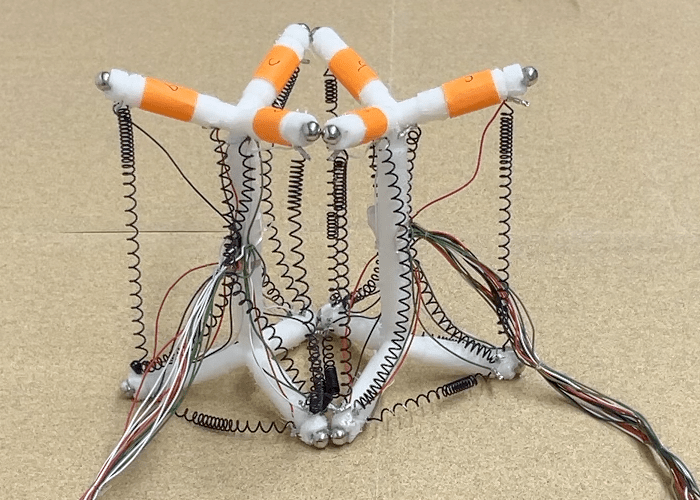}
  \caption{Fully attached}
  \label{fig:a3}
\end{subfigure}%
\caption{Attachment between single modules}
 \vspace{-1\baselineskip}
\label{fig:single_module_attachments}
\end{figure}

Once two modules are in close proximity they enter an attachment protocol to ensure a fully connected four-magnet bond (Fig.~\ref{fig:single_module_attachments}). During this protocol, one module actively moves towards the other one while the other module\secondversion{, while the second module}waits and cooperates with the active module. Once two modules are close enough, the head of the passive module leans outward to prevent the upper magnets from attaching together, since an initial attachment of the top magnetic balls can prevent successful attachment of the bottom two balls (because the modules are unable to shift their center of gravity sufficiently to slide the bottom feet).

When at least one pair of bottom feet has been attached as shown in Fig.~\ref{fig:a2}, all the SMA actuators on the two modules are commanded to release until returning to initial states. During this process,\secondversion{all}other pairs of magnetic balls will\secondversion{typically}be attached automatically as shown in Fig.~\ref{fig:a3}. For those exceptional cases where \hlsecondversion{not all}the remaining magnetic balls are\secondversion{not}attached automatically, the strategy we used is to contract the four neighboring vertical SMAs simultaneously and then contract the two front horizontal SMAs at the same time.\secondversion{if needed} 

\subsection{\hlsecondversion{Attachment Test}}

To test this attachment strategy we conducted an experiment where we attached one of the bottom magnets and varied the alignment angle between the modules (Fig.~\ref{fig:a2} is an example with an approximately $10\degree$ alignment angle). We tested 5 trials for each angle ($0\degree$, $30\degree$, $60\degree$, $90\degree$, $120\degree$). The success rate was 100\% for angles less than or equal to 90 degrees, while 4 out of 5 trials with $120\degree$ succeeded. Furthermore, for trials with angle $0\degree$, $30\degree$, and $60\degree$, the remaining pair of magnetic balls automatically attached during recovery process. For trials with angle $90\degree$, 2 trials out of 5 required an additional approach after recovery to help with attachment.

% all other pairs of magnetic balls will be attached automatically as shown in Fig.~\ref{fig:a3} unless the angle between two modules is too large (\hl{what is the maximum angle?}). Thus, in order to successfully attach two modules, it is necessary to align the direction of two modules at first.

%%% 3-module locomotion figure
\begin{figure*}[h]
\begin{center}
\begin{subfigure}{0.2076\textwidth}
\includegraphics[width=\linewidth]{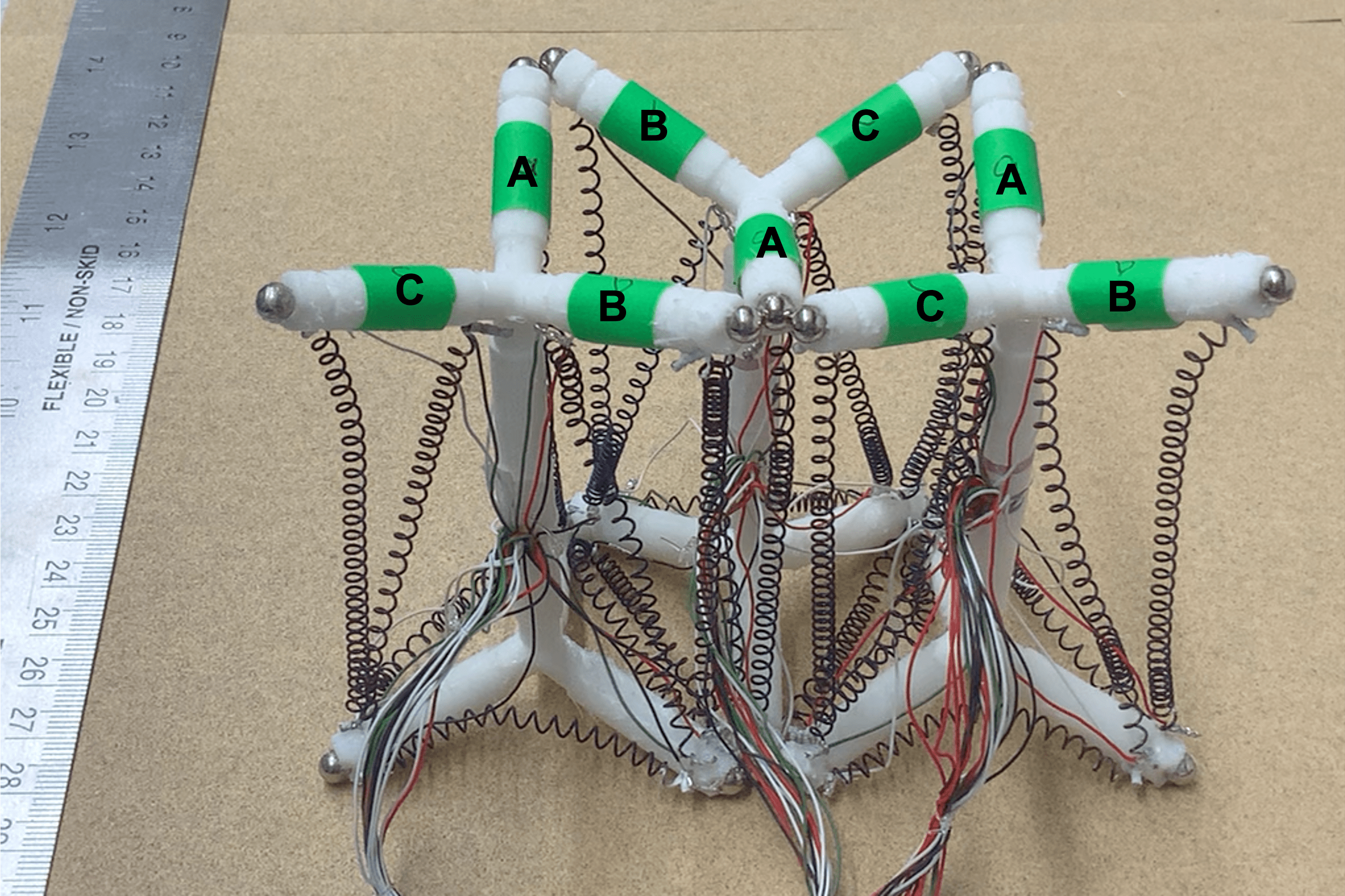}
  \caption{Initial state} \label{fig:3-module_initial}
\end{subfigure}%
\begin{subfigure}{0.171\textwidth}
  \includegraphics[width=\linewidth]{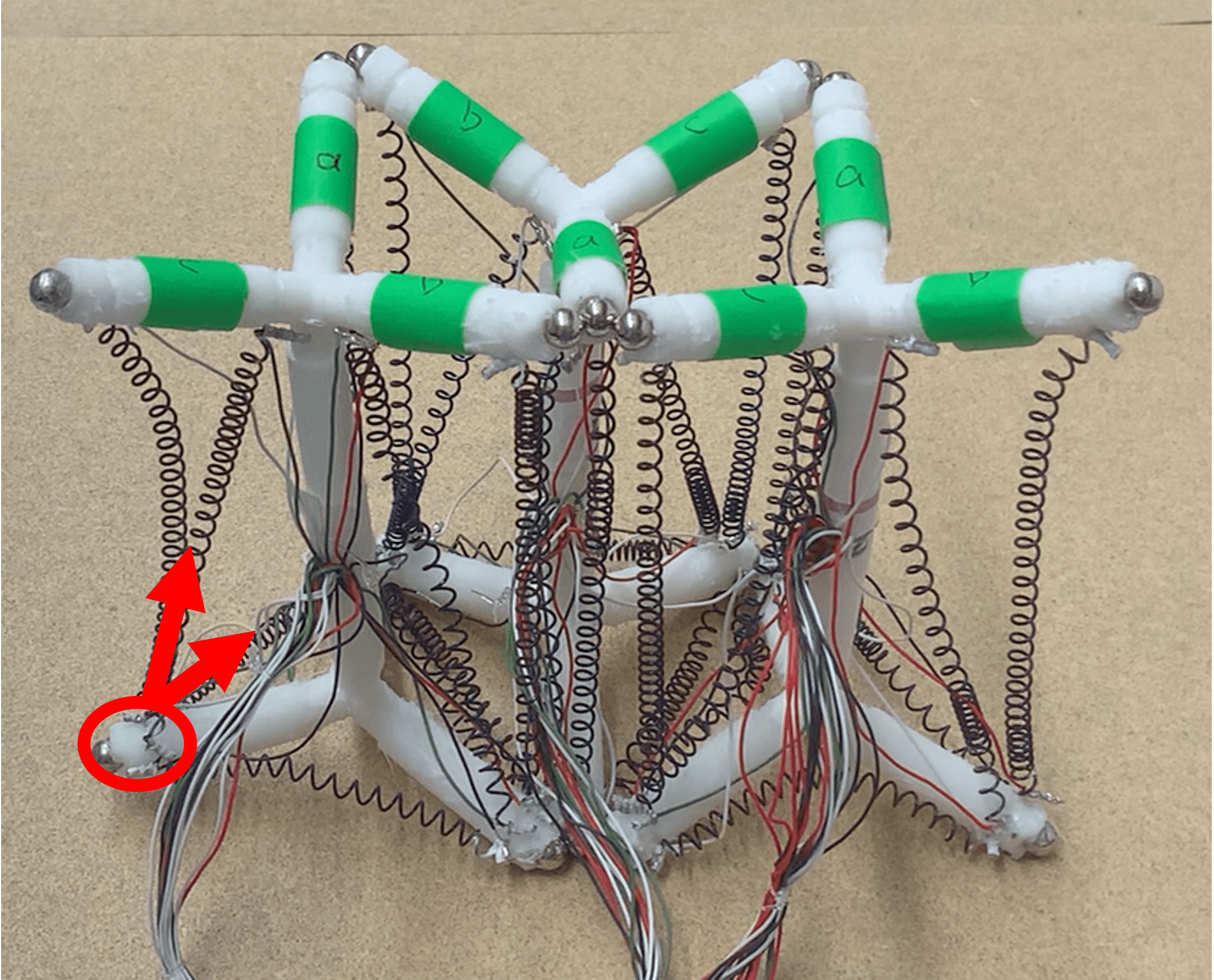}
  \caption{Left foot forward} \label{fig:l-2}
\end{subfigure}%
\begin{subfigure}{0.171\textwidth}
  \includegraphics[width=\linewidth]{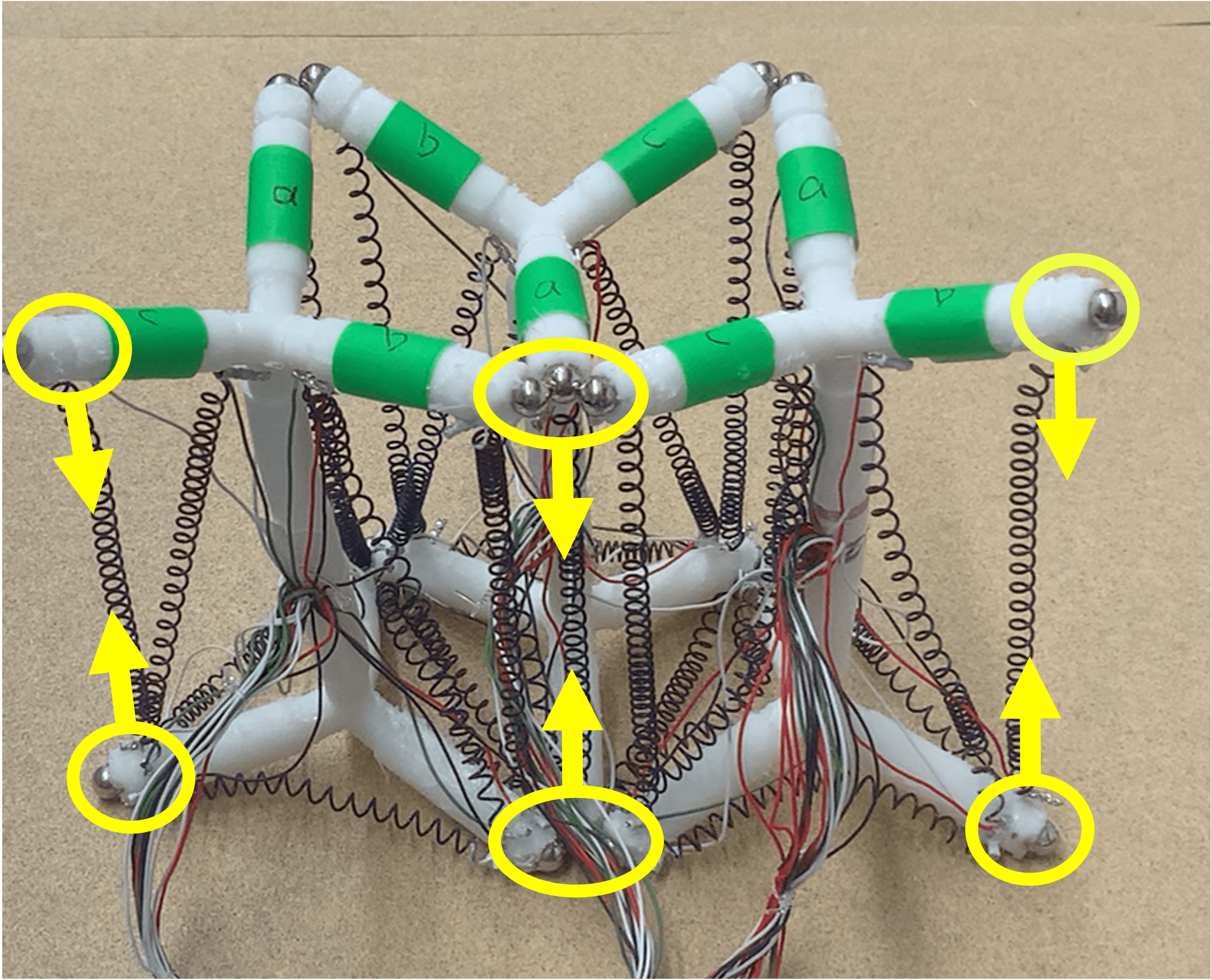}
  \caption{Leaning back} \label{fig:l-3}
\end{subfigure}%
\begin{subfigure}{0.171\textwidth}
  \includegraphics[width=\linewidth]{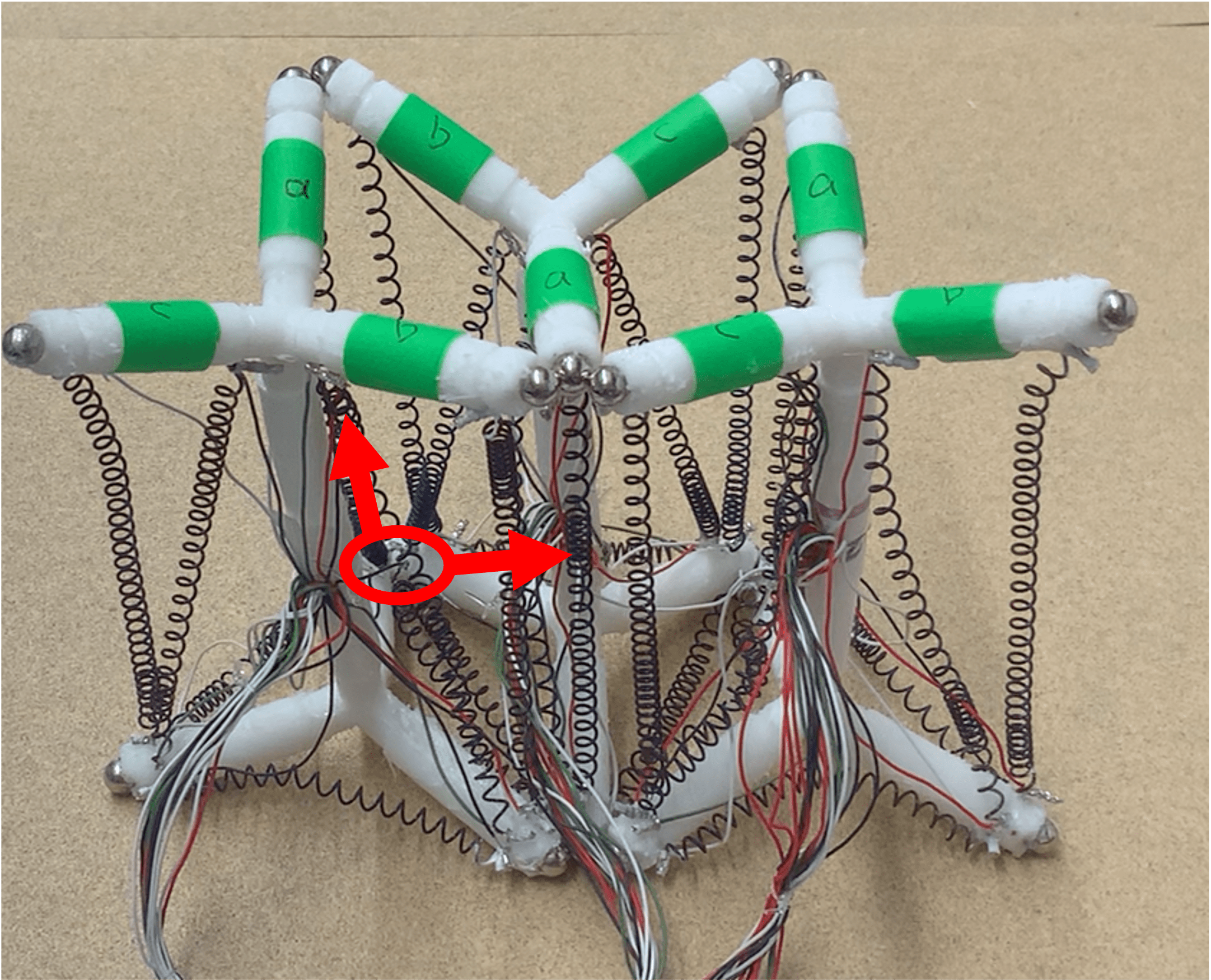}
  \caption{
  %Phase 2: 
  Front contracted} \label{fig:l-4}
\end{subfigure}%
\begin{subfigure}{0.171\textwidth}
  \includegraphics[width=\linewidth]{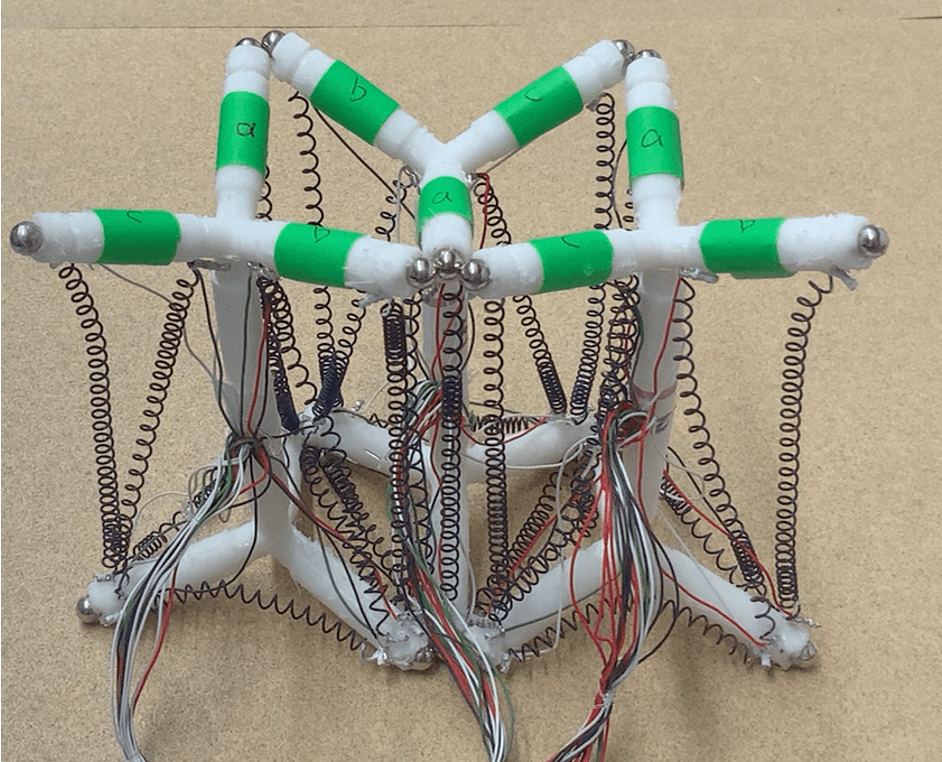}
  \caption{
  %Phase 2: 
  Release} \label{fig:l-5}
\end{subfigure}
\caption{A sequence of actions the 3-module takes to move the left feet forward}
\label{fig:3-module_unit_loco}
\vspace{-1\baselineskip}
\end{center}
\end{figure*}
\section{3-module Locomotion \& Attachment}
\label{sec:3-loco}
The locomotion pattern of 3-module is like a three-legged race, where the feet in between are strapped by magnetic balls and only the left most and right most feet can move freely. 
%They are cooperating together to move forward.

\begin{figure}[t]
\centering
\begin{subfigure}{.3\linewidth}
  \centering
  \includegraphics[width=\linewidth]{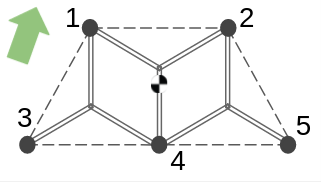}
  \caption{}
  \label{fig:3-module_unit_physical: a}
\end{subfigure}
\begin{subfigure}{.3\linewidth}
  \centering
  \includegraphics[width=\linewidth]{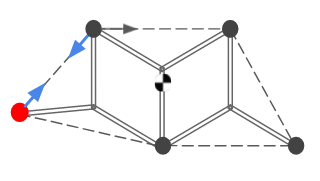}
  \caption{}
  \label{fig:3-module_unit_physical: b}
\end{subfigure}
\begin{subfigure}{.3\linewidth}
  \centering
  \includegraphics[width=\linewidth]{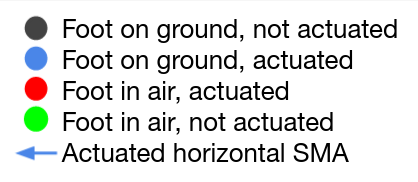}
  \vspace{0.5cm}
\end{subfigure}
\begin{subfigure}{.3\linewidth}
  \centering
  \includegraphics[width=\linewidth]{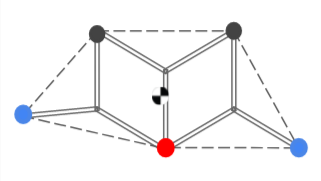}
  \caption{}
  \label{fig:3-module_unit_physical: c}
\end{subfigure}
\begin{subfigure}{.3\linewidth}
  \centering
  \includegraphics[width=\linewidth]{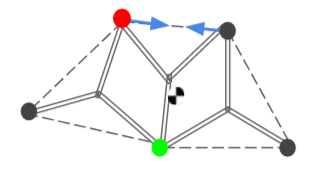}
  \caption{}
  \label{fig:3-module_unit_physical: d}
\end{subfigure}
\begin{subfigure}{.3\linewidth}
  \centering
  \includegraphics[width=\linewidth]{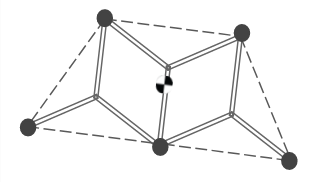}
  \caption{}
  \label{fig:3-module_unit_physical: e}
\end{subfigure}
\caption{\centering Physical analysis of three-module locomotion \newline
(left-leg step, top-down view of bottom legs)}
\vspace{-1\baselineskip}
\label{fig:3-module_unit_physical}
\end{figure}

\begin{figure}[htb]
\centering
% \vspace*{-3mm}
  \includegraphics[width=0.6\linewidth]{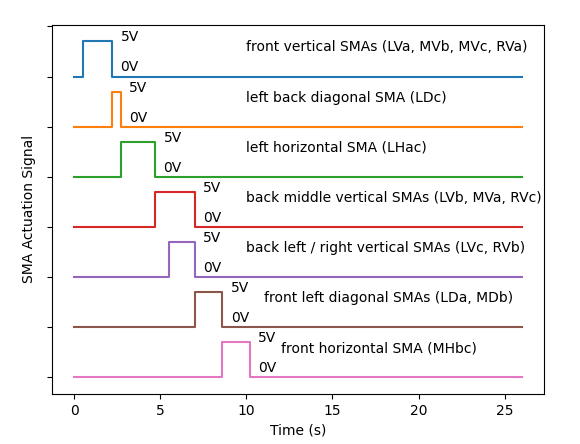}
\caption{Control signal of three-module locomotion}
 \vspace{-1\baselineskip}
\label{fig:three_module_control_signal}
\end{figure}

\subsection{Physical Analysis of ``Three-legged Race" Pattern\secondversion{Locomotion}}
The principle of moving forward for 3-module unit is similar to the combined {\em Shuffling} and {\em Grab-and-pull} pattern of single module locomotion, where left foot and right foot are alternately moving forward, and the front SMA is actuated to speed up the forward movements. The main difference and advantage of three-module locomotion is that given the greater number of ground contact points, feet can be lifted and advanced in the air while maintaining stability, thereby eliminating sliding friction and enabling significantly faster locomotion as compared to a single module.

%, where there is no contact with the ground and no friction force when moving forward, therefore, its speed is much faster than the ``Shuffle" pattern.

Take the left foot as an example.\secondversion{As Fig.{~\ref{fig:3-module_unit_physical: b}} shows, }The first step is to raise the left foot and move it forward in the air \hlsecondversion{(Fig.{~\ref{fig:3-module_unit_physical: b}}).}The second step (Fig.~\ref{fig:3-module_unit_physical: c}) plants the outside back feet (nodes 3 \& 5) by switching the center of gravity backward while lifting and advancing the center-back leg (node 4). The last step is to raise the front left foot (node 1) and actuate the front horizontal SMA to reduce the friction that influences the movement. As shown in Fig.~\ref{fig:3-module_unit_physical: d} only three feet remain on the ground. Finally, the three-module can move the left side of the body forward\secondversion{as shown in}(Fig.~\ref{fig:3-module_unit_physical: e}). The sequence will then be mirrored for the right side. Photos of the gait sequence\secondversion{can be seen}\hlsecondversion{shown}in Fig.~\ref{fig:3-module_unit_loco} and Fig.~\ref{fig:three_module_control_signal} shows the control signal for moving the left foot forward (Lx, Mx, Rx represent the SMAs on the left, middle and right modules separately, with x labelled in the same way as in Fig.~\ref{fig:system_overview_singlemodule}, with limb labels (A, B, C) for each module shown in Fig.~\ref{fig:3-module_initial}).

\subsection{Experiments}

% To actuate the modules, the front vertical SMAs will be actuated after the initial state as Fig.~\ref{fig:3-module_initial} shown (corresponding to Fig.~\ref{fig:3-module_unit_physical: a}) for speeding up the recovery process of the whole structure as only back SMAs are required to contract (Fig.~\ref{fig:3-module_unit_physical: c}) will lead the whole skeleton of the structure leaning back which requires a long time to fully recover. Deploying an external force in the front helps speed up the recovery.

% Fig.~\ref{fig:l-1} and Fig.~\ref{fig:l-2} shows the actions where the left foot is lifted and is moving forward as Fig.~\ref{fig:3-module_unit_physical: b} analyzed.  Fig.~\ref{fig:l-3} and Fig.~\ref{fig:l-4} relates to Fig.~\ref{fig:3-module_unit_physical: c} and Fig.~\ref{fig:3-module_unit_physical: d}, where all back vertical SMA are contracted, left front foot is lifted, and front horizontal SMA has been contracted. Fig.~\ref{fig:3-module_unit_physical: e} shows the state after recovery.

To move forward for $2~cm$, the running time of 3-module unit was $2.35~min$ (median of 5 trials) and the cooling time for each cycle was $12~s$ out of $22.6~s$, or $0.0335~BL/s$. This motion is faster than a single module as each foot can be lifted and then moved forward in air, allowing more forward displacement during each cycle. 

\subsection{Attachment Between 3-module Units}

% \begin{figure}[htb]
% \centering
% \begin{subfigure}{.24\linewidth}
%   \centering
%   \includegraphics[width=\linewidth]{images/3attach.001.jpeg}
%   \caption{}
%   \label{fig:3-module_unit_attachments:a}
% \end{subfigure}
% \begin{subfigure}{.24\linewidth}
%   \centering
%   \includegraphics[width=\linewidth]{images/3attach.002.jpeg}
%   \caption{}
%   \label{fig:3-module_unit_attachments:b}
% \end{subfigure}
% % \begin{subfigure}{.15\linewidth}
% %   \centering
% %   \includegraphics[width=\linewidth]{images/3attach.003.jpeg}
% % \end{subfigure}
% % \begin{subfigure}{.15\linewidth}
% %   \centering
% %   \includegraphics[width=\linewidth]{images/3attach.004.jpeg}
% % \end{subfigure}
% \begin{subfigure}{.24\linewidth}
%   \centering
%   \includegraphics[width=\linewidth]{images/3attach.005.jpeg}
%   \caption{}
%   \label{fig:3-module_unit_attachments:c}
% \end{subfigure}
% \begin{subfigure}{.24\linewidth}
%   \centering
%   \includegraphics[width=\linewidth]{images/3attach.006.jpeg}
%   \caption{}
%   \label{fig:3-module_unit_attachments:d}
% \end{subfigure}
% \caption{Lattice self-assembly}
% %  \vspace{-1\baselineskip}
% \label{fig:3-module_unit_attachments}
% \end{figure}

Fig.~\ref{fig:system_overview_lattice_assembly} shows an example of attaching four three-module groups together. Unlike single-module attachment, the passive module unit waits motionlessly instead of leaning backward to cooperate, due to the complexity that would be required for the leaning motion. Moreover, since the 
% front vertical SMAs of the three-module unit is used for balancing rather of switching the center of gravity, the 
unit will not lean forward as the single module does during locomotion, either the bottom magnetic balls will be attached first when unit is leaning backward or all balls (top and bottom) will attach simultaneously when the unit is upright. 

When a 3-module unit approaches the lattice it is common for all magnetic connection points to snap together nearly simultaneously, but this is not always the case. Fig.~\ref{fig:c3} shows a case where only a partial connection is made between the orange unit and the lattice. In this case, we used a rotation motion to pivot the unit around the connected points until a full connection was reached. Closing the loop with sensing and fully automating this process is a next step in future work.

% Given there are two group of magnetic balls of the active unit that need to be attached with the passive lattice, where each group includes top balls and the corresponding bottom balls. When one group of magnetic balls(the top and bottom balls) are attached (Orange unit in Fig.\ref{fig:3-module_unit_attachments:c} shows an example), the active module will rotate to attach. The rotate function is similar to the locomotion of left/right foot, but the only difference is that the vertical SMAs near the attached nodes will not be actuated.

% \input{sections/design-3d.tex}
\section{Non-prehensile Manipulation}
\label{sec:manipulation}

In this section, we describe strategies to roll a single ball or multiple balls along the surface of the structure. Ultimately, we imagine transporting other objects or even modules themselves along the surface, using lattice deformation as a whole-body manipulation strategy.
% Our main goal is to implement the manipulation using assembled modules in 3D, and to illustrate easier, let's dive into a 2D example firstly where each module is a 3-bar tensegrity actuated by motors shown in Figure~\ref{fig:mani_2d}. By changing the distance between the top left node and the bottom node of each tensegrity structure, the top surface will change to minimum the potential energy of the whole system and it will act like a wave, which will induce the ball to move from left to right.

% \begin{figure}[h]
%     \centering
%     \includegraphics[width=0.2\linewidth]{images/manipulation_2d.png}
%     \caption{Manipulation with motors in 2D}
%     \label{fig:mani_2d}
% \end{figure}

\subsection{Bi-directional Linear Manipulation with SMAs in 3D}

% \begin{figure}[h]
% \centering
% \begin{subfigure}{.199\linewidth}
%   \centering
%   \includegraphics[width=\linewidth]{images/m2-new.jpeg}
% \end{subfigure}%
% \begin{subfigure}{.199\linewidth}
%   \centering
%   \includegraphics[width=\linewidth]{images/m3-new.jpeg}
% \end{subfigure}%
% \begin{subfigure}{.199\linewidth}
%   \centering
%   \includegraphics[width=\linewidth]{images/m4-new.jpeg}
% \end{subfigure}%
% \begin{subfigure}{.199\linewidth}
%   \centering
%   \includegraphics[width=\linewidth]{images/m5-new.jpeg}
% \end{subfigure}%
% \begin{subfigure}{.199\linewidth}
%   \centering
%   \includegraphics[width=\linewidth]{images/m6-new.jpeg}
% \end{subfigure}%
% \caption{Bi-directional manipulation by self-assembled lattice}
% \label{fig:mani_3d}
% \end{figure}

\secondversion{To manipulate the ball,}Vertical SMAs are actuated to change the shape of the top surface to make the ball roll. Fig.~\ref{fig:bi_complex} shows an example of bi-direction linear manipulation.\secondversion{using vertical SMAs.}

\begin{figure}[h]
\centering
\begin{subfigure}{.4\linewidth}
  \centering
  \includegraphics[height=2.6cm]{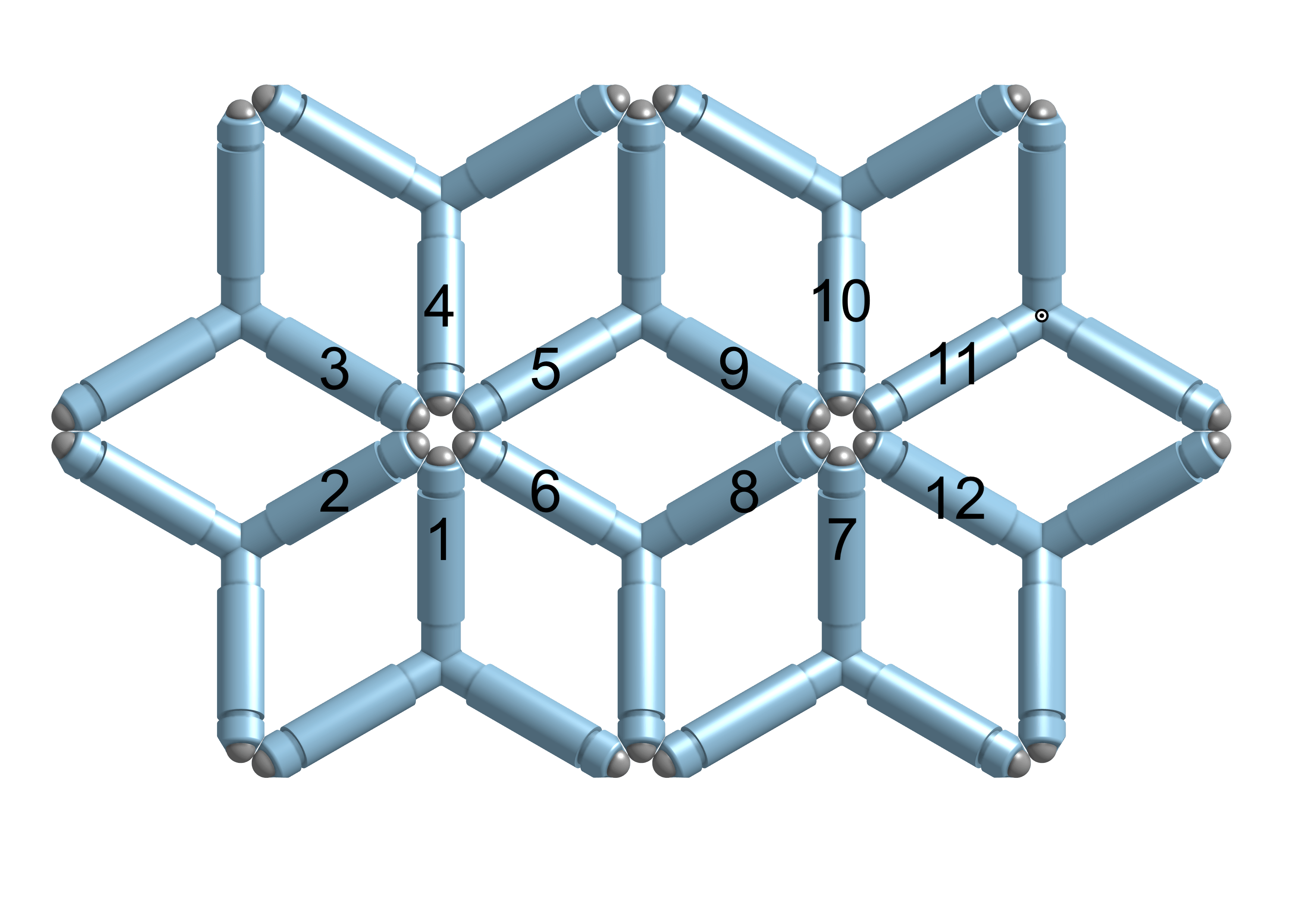}
  \caption{Top view}
  \label{fig:mani_top_view}
\end{subfigure}%
\begin{subfigure}{.4\linewidth}
  \centering
  \includegraphics[height=2.5cm]{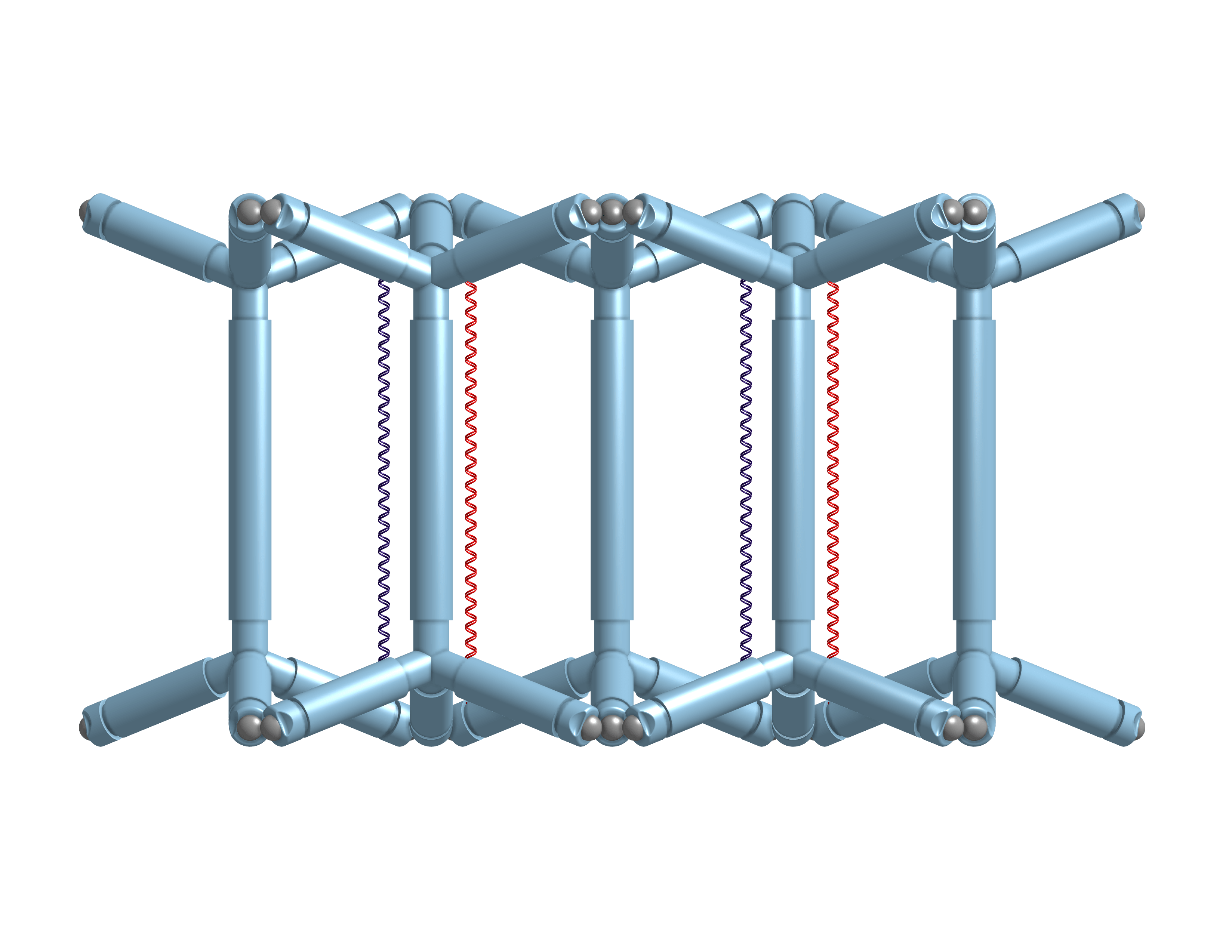}
  \caption{Front view}
  \label{fig:mani_front_view}
\end{subfigure}
\caption{Top and front view of lattice with 10 modules.}
 \vspace{-1\baselineskip}
\label{fig:mani_cad}
\end{figure}

\begin{figure}[h]
\centering
\begin{subfigure}{.4\linewidth}
  \centering
  \includegraphics[width=.9\linewidth]{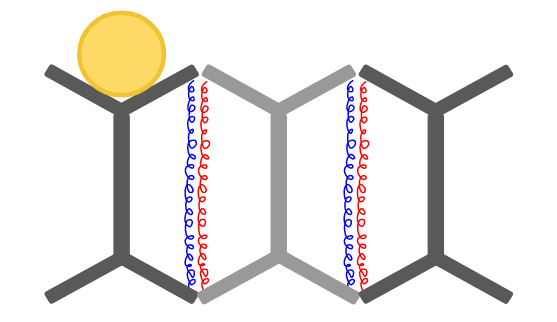}
  \caption{Initial state}
  \label{fig:mani_initial_with_feet}
\end{subfigure}%
\begin{subfigure}{.4\linewidth}
  \centering
  \includegraphics[width=.9\linewidth]{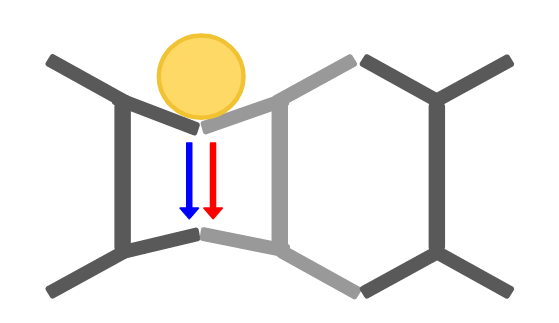}
  \caption{Actuate left node SMAs}
  \label{fig:mani_first_step_with_feet}
\end{subfigure}
\bigskip
\begin{subfigure}{.4\linewidth}
  \centering
  \includegraphics[width=.9\linewidth]{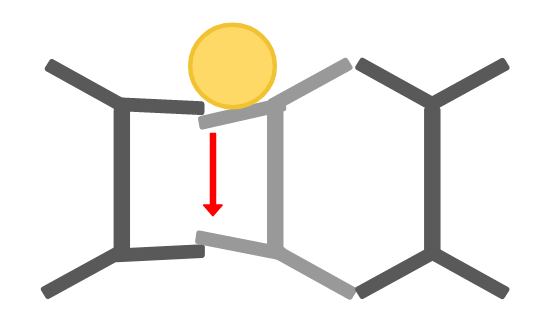}
  \caption{Actuate left connector}
  \label{fig:mani_second_step_with_feet}
\end{subfigure}%
\begin{subfigure}{.4\linewidth}
  \centering
  \includegraphics[width=.9\linewidth]{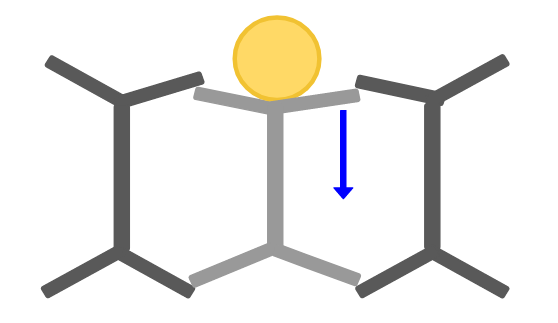}
  \caption{Actuate right connector}
  \label{fig:mani_third_step_with_feet}
\end{subfigure}%
 \vspace{-1\baselineskip}
\caption{Manipulation process}
 \vspace{-1\baselineskip}
\label{fig:mani_process_with_feet}
\end{figure}

\begin{figure}[h]
\centering
 \vspace{-1\baselineskip}
\includegraphics[width=0.7\linewidth]{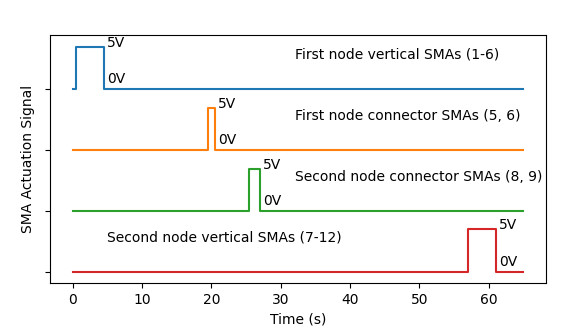}
\caption{Control signal of manipulation \hlfinalversion{(1st node $\rightarrow$ 2nd node)}\finalversion{from the first node to the second node}}
\label{fig:manipu_control_signal}
 \vspace{-1\baselineskip}
\end{figure}

An example of moving a ball from the left node to the right node is shown in Fig.~\ref{fig:mani_process_with_feet}. The first step is to drag the left node down vertically by actuating six connected vertical SMAs\secondversion{connected to the left node}(shown in Fig.~\ref{fig:mani_first_step_with_feet}). During release of these SMAs, the height of left node increases gradually. 
When the left node rises, the ball might drop either left and right; to avoid this, before the left node becomes flat, the first node connector SMAs (marked in red in Fig.~\ref{fig:mani_second_step_with_feet}) is actuated to give the ball velocity in the desired direction. After 5 seconds, the second node connector SMAs (marked in blue in Fig.~\ref{fig:mani_third_step_with_feet}) is actuated to give the ball a rightwards velocity\secondversion{to move further to the right and end up at}towards a stable position on top of the middle module.
The corresponding control signal is shown in Fig.~\ref{fig:manipu_control_signal}  (vertical actuated SMAs are labeled in Fig.~\ref{fig:mani_top_view}).

\subsection{Path Control of a Ball }
% \begin{figure*}[h]
% \centering
% \begin{subfigure}{.25\linewidth}
%   \centering
%   \includegraphics[width=\linewidth]{images/twoballs1-min.png}
% \end{subfigure}%
% \begin{subfigure}{.25\linewidth}
%   \centering
%   \includegraphics[width=\linewidth]{images/twoballs2-min.png}
% \end{subfigure}%
% \begin{subfigure}{.25\linewidth}
%   \centering
%   \includegraphics[width= \linewidth]{images/twoballs3-min.png}
% \end{subfigure}%
% \begin{subfigure}{.25\linewidth}
%   \centering
%   \includegraphics[width= \linewidth]{images/twoballs4-min.png}
% \end{subfigure}%
% \caption{Example of manipulating two balls simultaneously}
% \label{fig:mani_3d_two_balls}
% \end{figure*}

\begin{figure}[h]
\centering
\begin{subfigure}{0.33\linewidth}
    \centering
    \includegraphics[width=\linewidth]{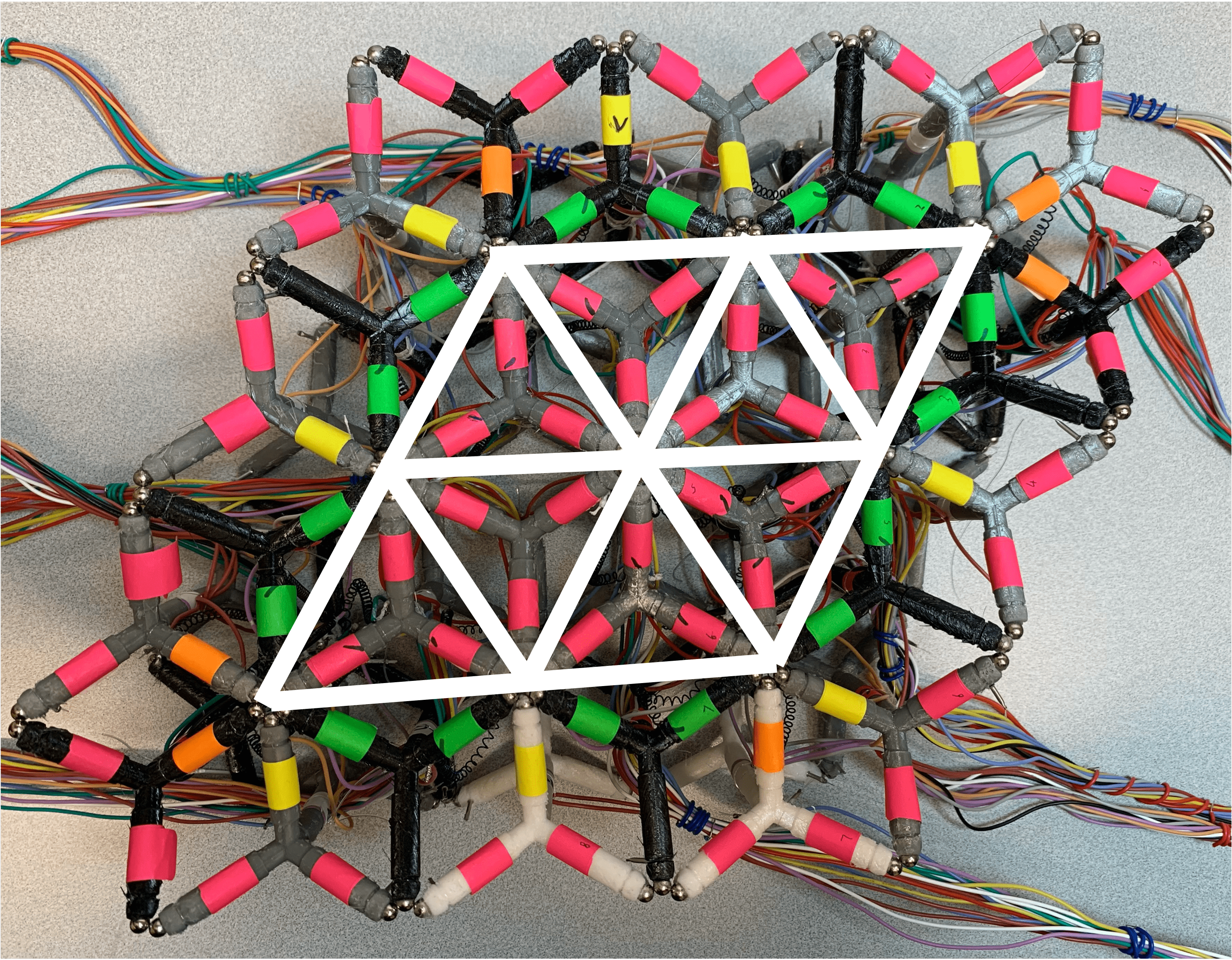}
    \caption{}
    \label{fig:path_control}
\end{subfigure}
\begin{subfigure}{.3\linewidth}
 \centering
    \includegraphics[width=\linewidth]{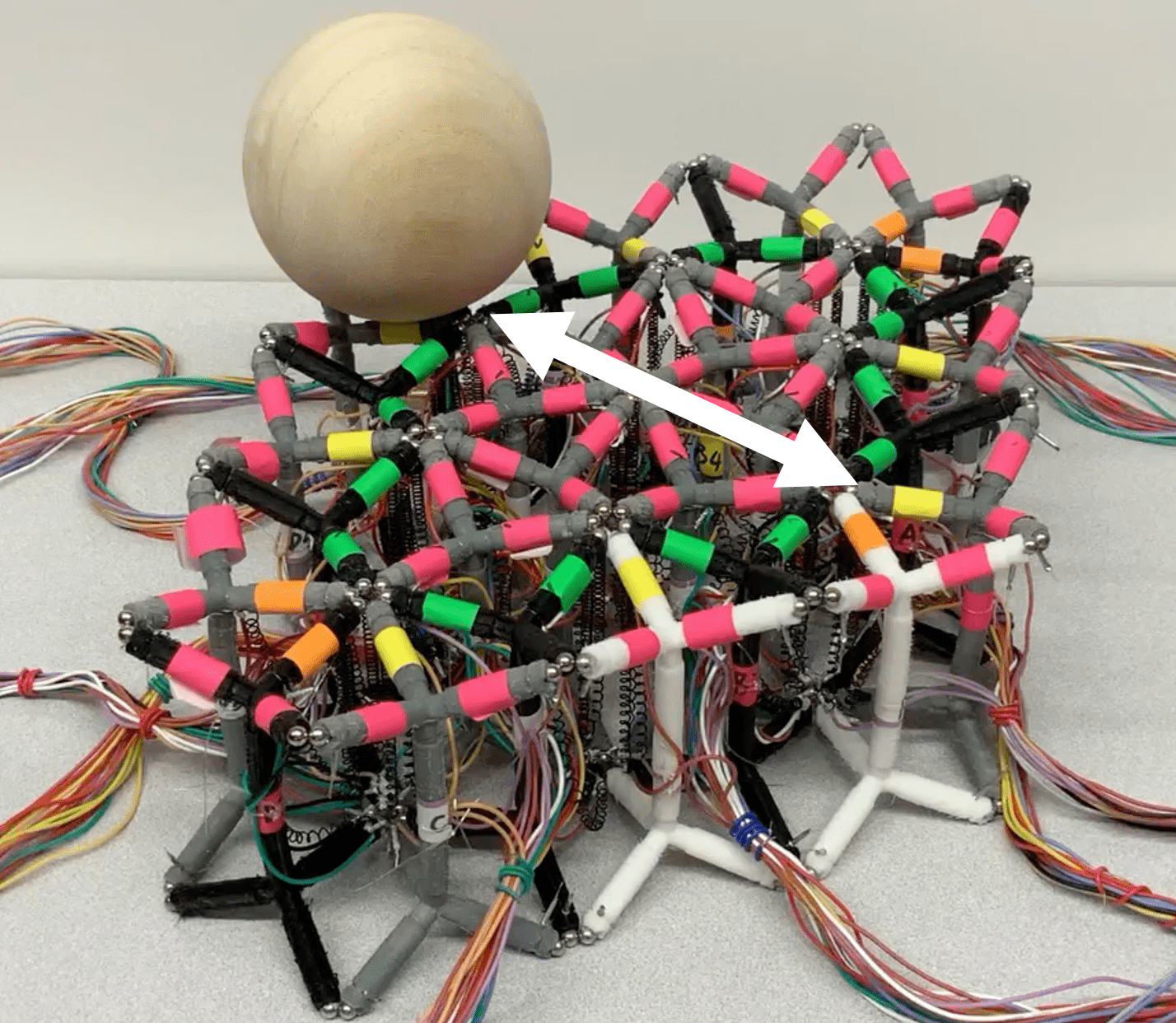} 
    \caption{}
    \label{fig:bi_complex}
\end{subfigure}
\begin{subfigure}{.33\linewidth}
    \centering
    \includegraphics[width=\linewidth]{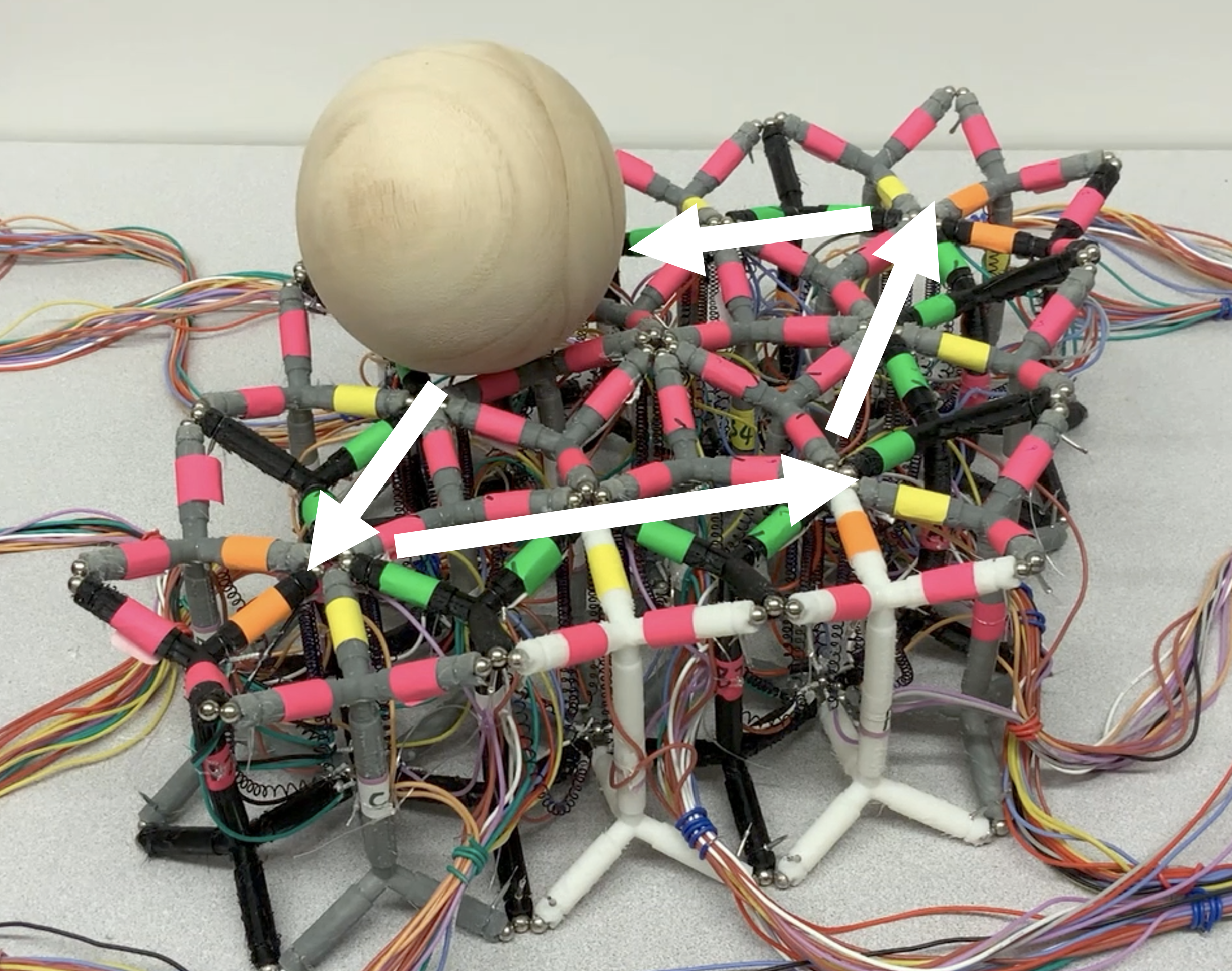} 
    \caption{}
    \label{fig:para_complex}
\end{subfigure}%
\caption{(a) Path control of a ball (b) Bi-direction manipulation (c) Parallelogram manipulation}
\vspace{-1\baselineskip}
\label{fig:complex_manipulations}
\end{figure}

Using the same principle, adding additional modules to the lattice allows us to do path control of a ball along the top surface. Fig.~\ref{fig:path_control} shows an example of a 3x3 grid, where white edges are the allowed paths to move through. 
\hlsecondversion{Except bi-directional manipulation,}we \hlsecondversion{also}tested\secondversion{bi-directional manipulation and}parallelogram manipulation on this surface\secondversion{as shown in Fig.{~\ref{fig:bi_complex}} and}(Fig.~\ref{fig:para_complex})\secondversion{We also tested}\hlsecondversion{and}manipulating two balls on the top surface, where as Fig.~\ref{fig:system_overview_manipulation} shows, two balls move in opposite directions without touching each other.

%\secondversion{
%\subsection{Multi-object Manipulation}
% subsection of 1 sentence is a bit too short! -djb

\section{Gripper formed from modules}
\label{sec:gripper}

% \begin{figure}[h]
%     \centering
%     \includegraphics[width=0.8\linewidth]{images/gripper1.png}
%     \caption{Various objects grasped by module-based gripper}
%     \label{fig:gripper}
% \end{figure}

% \hl{\textbf{Julien comment: We need to expand this section. Suggest splitting it into two paragraphs and an additional figure. We need to describe:}}
% \begin{enumerate}
%     \item \hl{the gripper configuration (how the modules are arranged \& why). There should be an illustration of the arrangement. Can briefly describe the method of performing the gripping (i.e. which wires are actuated).}
%     \item \hl{The test setup. What objects were tested, and what were the success rates of each type of object. Then, discuss the results. How does weight/shape/size/surface of object affect gripper success? Is it currently useful for applications or what improvements should be made?}
% \end{enumerate}

\hlsecondversion{We have also tested using a module-based gripper to grasp various objects, as shown in Fig.~{\ref{fig:system_overview_gripper}}. Top horizontal SMAs are added for each module such that the object can be grasped when both top and bottom horizontal SMAs on the inside boundary are actuated. To grasp larger objects, top and bottom horizontal SMAs on the outside boundary are actuated to open the gripper first. To avoid detachments between modules easily when grasping heavy objects, we used larger ball magnets with 5/16" diameter and $2.52$ lbs pull force.}

\hlsecondversion{Objects with different weight/shape/size/surface as shown in Fig.~{\ref{fig:system_overview_gripper}} were tested, and the results are shown in Table~{\ref{tab:gripping_results}}.
The gripper appears capable of gripping a variety of objects with a success rate that varies depending on the roughness of the gripper and the weight/shape/size/surface of the object.
To increase the success rate, future work could test increasing the roughness of the gripper, using magnets with higher pulling force, using softer skeletons to allow more deformation, or trying new arrangements of modules to grasp different objects.}

\begin{table}[]
\centering
\caption{Gripping configuration success rates}
\begin{tabular}{|cccccc|}
\hline
\textbf{Fig.~\ref{fig:g1}} & \textbf{Fig.~\ref{fig:g2}} & \textbf{Fig.~\ref{fig:g3}} & \textbf{Fig.~\ref{fig:g4}} & \textbf{Fig.~\ref{fig:g5}} & \textbf{Fig.~\ref{fig:g6}} \\ 
5/5         & 5/5         & 5/5         & 4/5         & 4/5         & 3/5         \\ \hline
\end{tabular}
 \vspace{-1\baselineskip}
\label{tab:gripping_results}
\end{table}
\section{Conclusion and Future work}
\label{sec:conclusion}

In this paper, we instantiated soft lattice modules that behave both independently and collectively. Using these lattice modules, we demonstrated preliminary steps toward the self-assembly of robotic lattice structures and employed several behaviors to perform collective and coordinated tasks. 
%were able to demonstrate the first steps towards a self-assembling soft robot lattice, as well as a form of peristaltic manipulation atop the mesh surface. 
%However, as a first step towards a soft modular mesh,

Our system has several limitations and opportunities for further development. 
At present, all the modules are tethered to an Arduino which controls the lattice as a whole.\secondversion{SMAs are capable for a tether-less system, thus in a fully realized system, t}To allow modules be self-contained and autonomous,\secondversion{SMAs could still be applied while}on-board computing, power, and sensing would be required. The modules would need\secondversion{a method of}communication\secondversion{, whether}through a contact-based mesh scheme or a wireless scheme. Methods of realizing distributed control of the lattice would need to be implemented.
% a tether-less system is a goal for future work

% Our approach for single module locomotion relies on friction, and the speed is limited even on rough surfaces. However, for 3-module locomotion, since the feet can be lifted, the speed is much faster. 
% Other forms of locomotion, such as rolling or jumping, might be faster. We focus on walking and sliding motions as they provide the needed precision for docking, as well as multi-module locomotion. 
Our single module locomotion strategy relies on friction, so the speed is limited even on rough surfaces.
However, for 3-module locomotion, since the feet can be lifted, the speed is much faster.
We plan to explore moving more modules together for centipede-type locomotion, and improve module design to make it walk individually instead of sliding. 

% \hl{Our single module locomotion strategy relies on friction, so the speed is limited even on rough surfaces, but when the feet can be lifted in 3-module locomotion, the speed increases. Though other forms of locomotion, such as rolling or jumping, might be faster, considering a higher precision for docking and the possibility to do multi-module locomotion, we choose to focus on walking and sliding patterns. For the next step, we plan to explore moving more modules together for centipede-type locomotion, and to test an improved module design that is individually capable of performing a walking gait.}

% djb -- this isn't really future work:
%\hl{Moreover, SMAs slow the locomotion speed since cooling time is much larger than actuation time. Our module-based gripper demonstrates how SMAs can be 
%useful without compromising speed.}

% What are weaknesses and limitations?
% cannot detach
% As mentioned above, for this paper, we are at the first step towards using modular tensegrity robot for locomotion and manipulation, although we can achieve attachment between modules, for the phase of detachments, we are still at the exploring stage. One possible solution to that could be using the principle of ``tug of war". 

Our system is capable of self-assembly, but the magnets present a challenge for self-disassembly. One potential approach could use the principle of ``tug of war" to generate sufficient detachment force, but detaching multiple connection points simultaneously may be difficult and a hardware modification is perhaps necessary for this purpose.

% cannot manipulate other shapes
% Moreover, for manipulation, we have tested manipulate a ball with different sizes, but haven't tried other shapes. We would like to test moving other stuffs with various shapes.
% single locomotion relys on friction, so slow

% Another area of future development would be in 
% The primary contribution of this work is to explore a robotic system that can do peristaltic manipulation by attachments between flexible modules. 
Yet another direction for further development lies in simulation.\secondversion{Although we have not focused on simulation, }We have built a simple simulation using an energy-based quasi-static model that estimates the configuration with the minimal potential energy using Lagrange multipliers. Our\secondversion{current} model assumes a pseudo-rigid bodied skeleton and requires further development to better reflect the \hlsecondversion{system}dynamics\secondversion{of the system}.

% \begin{figure}[tbh]
%     \centering
%     \includegraphics[width=0.6\linewidth]{images/3d_tensegrity1.png}
%     \caption{Simulation for Manipulation}
%     \label{fig:mag_design}
% \end{figure}

% unbalance issue
% Since we haven't add sensor and perception into the system, we cannot control it well and there are some unbalance issue when locomoting, and we plan to add sensors and control it using some control technique, like using PID controller for the next step.
% SMA needs time to cool down
% Another factor that affects the speed of locomotion is that the cooling time for one-way SMA is relative long, one potential method to deal with this is to scale up the module and then use motors.

\printbibliography

@article{lee2017soft,
  title={Soft robot review},
  author={Lee, Chiwon and Kim, Myungjoon and Kim, Yoon Jae and Hong, Nhayoung and Ryu, Seungwan and Kim, H Jin and Kim, Sungwan},
  journal={International Journal of Control, Automation and Systems},
%   volume={15},
%   number={1},
%   pages={3--15},
  year={2017},
  publisher={Springer}
}

@inproceedings{buckner2021design,
  title={Design Parameters for Stacked-Ribbon Shape-Memory Alloy Bending Actuators},
  author={Buckner, Trevor L and Kramer-Bottiglio, Rebecca},
  booktitle={2021 IEEE 4th International Conference on Soft Robotics (RoboSoft)},
%   pages={579--582},
%   year={2021},
  organization={IEEE}
}

@article{zou2018reconfigurable,
  title={A reconfigurable omnidirectional soft robot based on caterpillar locomotion},
  author={Zou, Jun and Lin, Yangqiao and Ji, Chen and Yang, Huayong},
  journal={Soft robotics},
  volume={5},
  number={2},
%   pages={164--174},
  year={2018},
  publisher={Mary Ann Liebert, Inc. 140 Huguenot Street, 3rd Floor New Rochelle, NY 10801 USA}
}

@article{morin2014using,
  title={Using “Click-e-Bricks” to Make 3D Elastomeric Structures},
  author={Morin, Stephen A and Shevchenko, Yanina and Lessing, Joshua and Kwok, Sen Wai and Shepherd, Robert F and Stokes, Adam A and Whitesides, George M},
  journal={Advanced Materials},
  volume={26},
  number={34},
  pages={5991--5999},
  year={2014},
  publisher={Wiley Online Library}
}

@article{nemitz2016using,
  title={Using voice coils to actuate modular soft robots: wormbot, an example},
  author={Nemitz, Markus P and Mihaylov, Pavel and Barraclough, Thomas W and Ross, Dylan and Stokes, Adam A},
  journal={Soft robotics},
  volume={3},
  number={4},
  pages={198--204},
  year={2016},
  publisher={Mary Ann Liebert, Inc. 140 Huguenot Street, 3rd Floor New Rochelle, NY 10801 USA}
}

@article{booth2018omniskins,
  title={OmniSkins: Robotic skins that turn inanimate objects into multifunctional robots},
  author={Booth, Joran W and Shah, Dylan and Case, Jennifer C and White, Edward L and Yuen, Michelle C and Cyr-Choiniere, Olivier and Kramer-Bottiglio, Rebecca},
  journal={Science Robotics},
  volume={3},
  number={22},
  year={2018},
  publisher={Science Robotics}
}

@article{kwok2014magnetic,
  title={Magnetic assembly of soft robots with hard components},
  author={Kwok, Sen W and Morin, Stephen A and Mosadegh, Bobak and So, Ju-Hee and Shepherd, Robert F and Martinez, Ramses V and Smith, Barbara and Simeone, Felice C and Stokes, Adam A and Whitesides, George M},
  journal={Advanced Functional Materials},
  volume={24},
  number={15},
  pages={2180--2187},
  year={2014},
  publisher={Wiley Online Library}
}

@INPROCEEDINGS{Seok2010PeristalticLocomotion,  author={Seok, Sangok and Onal, Cagdas D. and Wood, Robert and Rus, Daniela and Kim, Sangbae},  booktitle={2010 IEEE International Conference on Robotics and Automation},   title={Peristaltic locomotion with antagonistic actuators in soft robotics},   
% year={2010},  
volume={},  
number={}  
% pages={1228-1233}
}

@article{
Omori2009PeristalticEarthworms,
author={Omori,Hayato and Nakamura,Taro and Yada,Takayuki},
year={2009},
title={An underground explorer robot based on peristaltic crawling of earthworms},
journal={The Industrial Robot},
volume={36},
number={4},
pages={358-364},
language={English}
}

@inbook{Parker2016,
	Address = {Cham},
	Author = {Parker, Lynne E. and Rus, Daniela and Sukhatme, Gaurav S.},
	Booktitle = {Springer Handbook of Robotics},
	Editor = {Siciliano, Bruno and Khatib, Oussama},
% 	Pages = {1335--1384},
	Publisher = {Springer International Publishing},
	Title = {Multiple Mobile Robot Systems},
	Year = {2016}}

@article{mondada2004swarm,
  title={SWARM-BOT: A new distributed robotic concept},
  author={Mondada, Francesco and Pettinaro, Giovanni C and Guignard, Andre and Kwee, Ivo W and Floreano, Dario and Deneubourg, Jean-Louis and Nolfi, Stefano and Gambardella, Luca Maria and Dorigo, Marco},
  journal={Autonomous robots},
  volume={17},
  number={2},
  pages={193--221},
  year={2004},
  publisher={Springer}
}

@ARTICLE{Murata2002M_TRAN,  author={Murata, S. and Yoshida, E. and Kamimura, A. and Kurokawa, H. and Tomita, K. and Kokaji, S.},  journal={IEEE/ASME Transactions on Mechatronics},   title={M-TRAN: self-reconfigurable modular robotic system},   year={2002},  volume={7},  number={4},  pages={431-441}}

@ARTICLE{Yim2002ModularRobots,  author={Yim, M. and Ying Zhang and Duff, D.},  journal={IEEE Spectrum},   title={Modular robots},   year={2002},  volume={39},  number={2},  pages={30-34}}

@INPROCEEDINGS{wei2010Sambot,  author={Wei, Hongxing and Cai, Yingpeng and Li, Haiyuan and Li, Dezhong and Wang, Tianmiao},  booktitle={2010 IEEE International Conference on Robotics and Automation},   title={Sambot: A self-assembly modular robot for swarm robot},   year={2010},  volume={},  number={}  
% pages={66-71}
}

@inproceedings{bruce2014superball,
  title={\ {SUPER}ball: Exploring tensegrities for planetary probes},
  author={Bruce, J. and Sabelhaus, A.P. and Chen, Y. and Lu, D. and Morse, K. and Milam, S. and Caluwaerts, K. and Agogino, A.M. and SunSpiral, V.},
  booktitle={12th International Symposium on Artificial Intelligence, Robotics and Automation in Space (i-SAIRAS)},
  year={2014}
}

@inproceedings{sabelhaus2015system,
  title={\ {S}ystem design and locomotion of {SUPER}ball, an untethered tensegrity robot},
  author={Sabelhaus, A.P. and Bruce, J. and Caluwaerts, K. and Manovi, P. and Firoozi, R.F. and Dobi, S. and Agogino, A.M. and SunSpiral, V.},
  booktitle={2015 IEEE International Conference on Robotics and Automation (ICRA)},
  pages={2867--2873},
  year={2015},
  organization={IEEE}
}

@inproceedings{he2020adaptable,
  title={\ {A}n Adaptable Robotic Snake Using a Compliant Actuated Tensegrity Structure for Locomotion},
  author={He, Q. and Post, M.A.},
  booktitle={Annual Conference Towards Autonomous Robotic Systems},
  pages={70--74},
  year={2020},
  organization={Springer}
}

@article{boxerbaum2012continuous,
  title={Continuous wave peristaltic motion in a robot},
  author={Boxerbaum, A.S. and Shaw, K.M. and Chiel, H.J. and Quinn, R.D.},
  journal={The International Journal of Robotics Research},
  volume={31},
  number={3},
  pages={302--318},
  year={2012},
  publisher={SAGE Publications Sage UK: London, England}
}

@article{donald2008self,
  title={\ {S}elf-organizing microrobots},
  author={Donald, B.R. and Gross, S. and others},
  journal={Advanced Manufacturing Technology},
  volume={29},
  number={7},
  pages={4--6},
  year={2008},
  publisher={Frost \& Sullivan}
}

@inproceedings{mas2012object,
  title={Object manipulation using cooperative mobile multi-robot systems},
  author={Mas, Ignacio and Kitts, Christopher},
  booktitle={Proceedings of the World Congress on Engineering and Computer Science},
  volume={1},
  year={2012}
}

@inproceedings{song2002potential,
  title={A potential field based approach to multi-robot manipulation},
  author={Song, Peng and Kumar, Vijay},
  booktitle={Proceedings 2002 IEEE International Conference on Robotics and Automation},
  volume={2},
  pages={1217--1222},
  year={2002},
  organization={IEEE}
}

@article{arai2002advances,
  title={Advances in multi-robot systems},
  author={Arai, Tamio and Pagello, Enrico and Parker, Lynne E and others},
  journal={IEEE Transactions on robotics and automation},
  volume={18},
  number={5},
  pages={655--661},
  year={2002},
  publisher={Citeseer}
}

@inproceedings{lessard2016bio,
  title={A bio-inspired tensegrity manipulator with multi-DOF, structurally compliant joints},
  author={Lessard, Steven and Castro, Dennis and Asper, William and Chopra, Shaurya Deep and Baltaxe-Admony, Leya Breanna and Teodorescu, Mircea and SunSpiral, Vytas and Agogino, Adrian},
  booktitle={2016 IEEE/RSJ International Conference on Intelligent Robots and Systems (IROS)},
%   pages={5515--5520},
  year={2016},
  organization={IEEE}
}

@inproceedings{bruce2014design,
  title={Design and evolution of a modular tensegrity robot platform},
  author={Bruce, Jonathan and Caluwaerts, Ken and Iscen, Atil and Sabelhaus, Andrew P and SunSpiral, Vytas},
  booktitle={2014 IEEE International Conference on Robotics and Automation},
%   pages={3483--3489},
%   year={2014},
%   organization={IEEE}
}

@inproceedings{skelton2001introduction,
  title={An introduction to the mechanics of tensegrity structures},
  author={Skelton, Robert E and Adhikari, Rajesh and Pinaud, J-P and Chan, Waileung and Helton, JW},
  booktitle={Proceedings of the 40th IEEE conference on decision and control},
  volume={5},
  pages={4254--4259},
  year={2001},
  organization={IEEE}
}

@inproceedings{Zwier2017MagneticSC,
  title={Magnetic Sphere Constructions},
  author={Henry Segerman and Rosa Zwier},
  year={2017}
}

@inproceedings{rubenstein2012kilobot,
  title={Kilobot: A low cost scalable robot system for collective behaviors},
  author={Rubenstein, Michael and Ahler, Christian and Nagpal, Radhika},
  booktitle={2012 IEEE international conference on robotics and automation},
%   pages={3293--3298},
  year={2012},
  organization={IEEE}
}

@article{CBalls_Chen_2019,
author = {Chen, Zhuxiang and Zhao, Chuanwu and Zhang, Yu and Zhu, Yanhe and Fan, Jizhuang and Zhao, Jie},
year = {2019},
month = {04},
% pages = {012006},
title = {C-Balls: A Modular Soft Robot Connected and Driven via Magnet Forced},
% volume = {1207},
journal = {Journal of Physics: Conference Series}
}

@inproceedings{ShapeBotsSuzukiEtAl,
author = {Suzuki, Ryo and Zheng, Clement and Kakehi, Yasuaki and Yeh, Tom and Do, Ellen Yi-Luen and Gross, Mark D. and Leithinger, Daniel},
title = {ShapeBots: Shape-Changing Swarm Robots},
year = {2019},
publisher = {Association for Computing Machinery},
% address = {New York, NY, USA},
booktitle = {Proceedings of the 32nd Annual ACM Symposium on User Interface Software and Technology},
% pages = {493–505},
numpages = {13},
keywords = {shape-changing user interfaces, swarm user interfaces},
% location = {New Orleans, LA, USA},
% series = {UIST '19}
}

@article{
SayedEtAlLimpetII,
author = {Sayed, Mohammed E. and Roberts, Jamie O. and McKenzie, Ross M. and Aracri, Simona and Buchoux, Anthony and Stokes, Adam A.},
title = {Limpet II: A Modular, Untethered Soft Robot},
journal = {Soft Robotics},
volume = {8},
number = {3},
pages = {319-339},
year = {2021}
}

@article{huang2020shape,
  title={Shape memory materials for electrically-powered soft machines},
  author={Huang, Xiaonan and Ford, Michael and Patterson, Zach J and Zarepoor, Masoud and Pan, Chengfeng and Majidi, Carmel},
  journal={Journal of Materials Chemistry B},
  volume={8},
  number={21},
  pages={4539--4551},
  year={2020},
  publisher={Royal Society of Chemistry}
}

@article{hajunlee,
author = {Hajun Lee  and Yeonwoo Jang  and Jun Kyu Choe  and Suwoo Lee  and Hyeonseo Song  and Jin Pyo Lee  and Nasreena Lone  and Jiyun Kim },
title = {3D-printed programmable tensegrity for soft robotics},
journal = {Science Robotics},
volume = {5},
number = {45},
% pages = {eaay9024},
year = {2020}
}

@INPROCEEDINGS{7759033,  
author={Tosun, Tarik and Davey, Jay and Liu, Chao and Yim, Mark},  
booktitle={2016 IEEE/RSJ International Conference on Intelligent Robots and Systems (IROS)},   
title={Design and characterization of the EP-Face connector},  year={2016},  
volume={},  
number={},  
pages={45-51}}

@article{Rieffel,
author = {Rieffel, John and Mouret, Jean-Baptiste},
year = {2018},
month = {04},
pages = {},
title = {Adaptive and Resilient Soft Tensegrity Robots},
volume = {5},
journal = {Soft Robotics}
}

@article{Li2019JelloCubeAC,
  title={JelloCube: A Continuously Jumping Robot With Soft Body},
  author={Shuguang Li and Daniela Rus},
  journal={IEEE/ASME Transactions on Mechatronics},
  year={2019},
  volume={24},
  pages={447-458}
}

@conference{yim,
author = {Mark Yim},
title = {A Reconfigurable Modular Robot with Many Modes of Locomotion},
booktitle = {Porc. of JSME Intl. Conf. on Advanced Mechatronics},
address = {Tokyo, Japan},
year = {1993}
}

@ARTICLE{8769941,
author={Liu, Chao and Whitzer, Michael and Yim, Mark},  journal={IEEE Robotics and Automation Letters},
title={A Distributed Reconfiguration Planning Algorithm for Modular Robots},
year={2019},
volume={4},
number={4},
pages={4231-4238}}

@inproceedings{spine-like,
author = {Mirletz, Brian and Park, In-Won and Flemons, Thomas and Agogino, Adrian and Quinn, Roger and Sunspiral, Vytas},
year = {2014},
month = {07},
pages = {},
title = {Design and Control of Modular Spine-Like Tensegrity Structures}
}

@INPROCEEDINGS{8967640,  
author={Liu, Chao and Yim, Mark},  
booktitle={2019 IEEE/RSJ International Conference on Intelligent Robots and Systems (IROS)},   title={Reconfiguration Motion Planning for Variable Topology Truss},   
year={2019},  
volume={},  
number={},  
pages={1941-1948}}

@article{isoperi,
author = {Nathan S. Usevitch  and Zachary M. Hammond  and Mac Schwager  and Allison M. Okamura  and Elliot W. Hawkes  and Sean Follmer },
title = {An untethered isoperimetric soft robot},
journal = {Science Robotics},
volume = {5},
number = {40},
% pages = {eaaz0492},
year = {2020}
}

@inproceedings{Zappetti2017BioinspiredTS,
  title={Bio-inspired Tensegrity Soft Modular Robots},
  author={D. Zappetti and Stefano Mintchev and Jun Shintake and Dario Floreano},
  booktitle={Living Machines},
  year={2017}
}

@ARTICLE{4141032,  
author={Yim, Mark and Shen, Wei-min and Salemi, Behnam and Rus, Daniela and Moll, Mark and Lipson, Hod and Klavins, Eric and Chirikjian, Gregory S.},  
journal={IEEE Robotics   Automation Magazine},   title={Modular Self-Reconfigurable Robot Systems [Grand Challenges of Robotics]},   year={2007},  volume={14},  number={1},  pages={43-52}}
\end{document}